\documentclass[submission,copyright,creativecommons]{eptcs}
\providecommand{\event}{First Workshop on Agents and Robots for Reliable Engineered Autonomy, 2020} % Name of the event you are submitting to
\usepackage{underscore}           % Only needed if you use pdflatex.
\usepackage{cite}
\usepackage{amsmath,amssymb,amsfonts}
\usepackage{algorithmic}
\usepackage{algorithm}
\usepackage{graphicx}
\usepackage{textcomp}
\usepackage{xcolor}
\usepackage{listings}
\usepackage{xcolor}
\usepackage{subfigure}

\definecolor{codegreen}{rgb}{0,0.6,0}
\definecolor{codegray}{rgb}{0.5,0.5,0.5}
\definecolor{codepurple}{rgb}{0.58,0,0.82}
\definecolor{backcolour}{rgb}{0.95,0.95,0.92}

\lstdefinestyle{mystyle}{
    backgroundcolor=\color{backcolour},   
    commentstyle=\color{codegreen},
    keywordstyle=\color{magenta},
    numberstyle=\tiny\color{codegray},
    stringstyle=\color{codepurple},
    basicstyle=\ttfamily\footnotesize,
    breakatwhitespace=false,         
    breaklines=true,                 
    captionpos=b,                    
    keepspaces=true,                 
    numbers=left,                    
    numbersep=5pt,                  
    showspaces=false,                
    showstringspaces=false,
    showtabs=false,                  
    tabsize=2
}

\def\BibTeX{{\rm B\kern-.05em{\sc i\kern-.025em b}\kern-.08em
    T\kern-.1667em\lower.7ex\hbox{E}\kern-.125emX}}

\title{Semi-supervised Learning From Demonstration Through Program Synthesis: An Inspection Robot Case Study
}

\author{Sim\'on C. Smith
\institute{Institute of Perception, Action and Behaviour\\School of Informatics\\
The University of Edinburgh\thanks{This work was supported by funding from the ORCA Hub EPSRC project (EP/R026173/1, 2017-2021).}\\
Edinburgh, United Kingdom}
\email{artificialsimon@ed.ac.uk}
\and
\qquad\qquad Subramanian Ramamoorthy
\institute{\qquad \qquad \qquad Institute of Perception, Action and Behaviour\\\qquad\qquad\qquad School of Informatics\\
\qquad\qquad\qquad The University of Edinburgh\\
\qquad\qquad\qquad Edinburgh, United Kingdom}
\email{\qquad\qquad\quad s.ramamoorthy@ed.ac.uk}
}
\def\titlerunning{Semi-supervised Learning From Demonstration Through Program Synthesis}
\def\authorrunning{S. C. Smith \& S. Ramamoorthy}

\begin{document}
\maketitle

\begin{abstract}
Semi-supervised learning improves the performance of supervised machine learning by leveraging methods from unsupervised learning to extract information not explicitly available in the labels. Through the design of a system that enables a robot to learn inspection strategies from a human operator, we present a hybrid semi-supervised system  capable of learning  interpretable and verifiable models from demonstrations. The system induces a controller program by learning from immersive demonstrations using sequential importance sampling. These visual servo controllers are parametrised by proportional gains and are visually verifiable through observation of the robot's position in the environment. Clustering and effective particle size filtering allows the system to discover  goals in the state space. These goals are used to label the original demonstration for end-to-end learning of behavioural models. The behavioural models are used for autonomous model predictive control and scrutinised for explanations. We implement causal sensitivity analysis to identify salient objects and generate counterfactual conditional explanations. These features enable decision making interpretation and post hoc discovery of the causes of a failure. The proposed system expands on previous approaches to program synthesis by incorporating repellers in the attribution prior of the sampling process. We successfully learn the hybrid system from an inspection scenario where an unmanned ground vehicle has to inspect, in a specific order, different areas of the environment. The system induces an interpretable computer program of the demonstration that can be synthesised to produce novel inspection behaviours. Importantly, the robot successfully runs the synthesised program on an unseen configuration of the environment while presenting explanations of its autonomous behaviour.
\end{abstract}

%\begin{IEEEkeywords}
%semi-supervised learning, learning from demonstration, program induction, program synthesis, explainability, counterfactual explanations, artificial intelligence, robotics
%\end{IEEEkeywords}

\section{Introduction}
In recent years, we have seen robots cross the chasms between laboratory testbeds and field deployments, increasingly involving operation alongside human co-workers. A particularly useful domain of application is that of robots that take over human tasks in hazardous environments such as offshore energy platforms ~\cite{hastie2018orca}. In contrast to more simple environments, such hazardous environment have a set of particular requirements for a robotic system to be successfully implemented. For example, to quickly respond to emergencies or to keep a vital part of the system running when a change in the environment has occurred. One must be able to rapidly program these robots to solve a range of tasks. One approach to rapidly reconfiguring robots for new tasks is to enable a human operator to teach the robot, by demonstrating the task and performing it rather than by explicitly writing programs. This is the approach of learning from demonstration (LfD). These task reconfigurations can be at the level of low-level motor commands or they may be at a higher level of abstraction as a combination of low-level tasks. When enough demonstration data is available, low-level tasks can be learned through non-linear regression techniques,
%including dynamic movement primitives~\cite{schaal2005learning} or Gaussian mixture regression~\cite{}.
like deep learning~\cite{bojarski2016end, fu2019neural}. These learning techniques present consistently higher performance when compared to other learning approaches~\cite{schaal2005learning, Gribovskaya:148817}. A downside is that such techniques have promoted performance advances at the expense of interpretability and flexibility~\cite{samek2017explainable, dovsilovic2018explainable}. Robotic systems trained with deep learning based LfD may be inscrutable to the typical operator, hence defeating the point of easy reconfigurability. In this sense, interpretability, flexibility and explainability are desiderata for the class of autonomous systems that will be deployed in such complex scenarios alongside humans.
%Our system  address these limitations by inferring high-level goals, inducing a program and training robust controllers from a limited set of demonstrations of the desired task.

To improve flexibility, compositional techniques can be used to divide demonstrations into sub-tasks. Learning composition of individual motions usually requires a set of primitives that are sequenced to solve more complex tasks~\cite{6386047, mangin:hal-00652346}. Usually, these approaches require the primitives to be known before learning~\cite{Billard:2013}. Approaches that automatically segment a demonstration~\cite{5650500, Kulic:2012:ILF:2159336.2159342} still have to deal with an unknown number of segments and task-dependent primitives.

We present a semi-supervised learning system that can induce a program from a demonstration, synthesise new programs and explain its behaviour. The unsupervised part of the system discovers proportional controllers that satisfy the demonstration.
%using particle filtering and by sampling from an attribution prior.
%By applying clustering techniques and effective particle size filters,
Our system can discover behavioural goals in the state space. The goals, visiting order and proportional controllers allow us to induce a computer program that abstracts the demonstration. This abstraction is a complexity reduction of the demonstration as it allows a user with programming skill to examine the full demonstration at once~\cite{lipton2018mythos}.

Continuing with the semi-supervised paradigm, we use the goals to label and segment the original demonstration. The newly labelled data is used to train end-to-end low-level behavioural model predictors.
These behavioural models have two main objectives. First, following a hybrid system of high- and low-level abstractions~\cite{lake2017building}, the behavioural predictions can be used to synthesise new programs that are verifiable by the robot even in unseen configurations of the environment. Second, using black-box analysis~\cite{samek2017explainable, selvaraju2017grad}, the behavioural models can explain the actual behaviour based on saliency in the input state of the end-to-end model and by counterfactual explanations. 

As mentioned before, this work builds on an existing probabilistic goal identification and program induction approach~\cite{Burke19Explanation}. However, this approach is particularly limited by the assumption that objects in a given scene are attractors to a robot. The likelihood of a position in an environment being a goal or target is assumed to be proportional to the saliency of the objects or regions in an image. The present work addresses this limitation by introducing repellers that allow the system to account for obstacles in the path of the robot. We extend the system by adding a likelihood term that includes the probability that a particle belongs to the path over the next steps of the robot. Results show that using this approach, our system can learn object avoidance behaviours without explicitly defining these as a high-level goal. As another extension to the original method, we are able to autonomously control the robot to follow a synthesised program containing behaviours unseen in the demonstrations. We test our system in an oil rig digital twin~\cite{pairet2019digital}, where an operator demonstrates an inspection task in an immersive teleoperation scenario. To summarise, our system can:
\begin{itemize}
  \item Automatically infer high-level goals in a surveillance task demonstrated by an operator
  \item Induce a computer program based on the demonstrations
  \item Automatic demonstration data labelling using unsupervised learning
  \item Learn behavioural models for predictive control (MPC) and explainability from the automatic labelled data
  \item Behaviour prediction, highlighting of prediction sensitive objects and counterfactual queries as post hoc explanations
\end{itemize}

\begin{figure}[!ht]
  \subfigure[System architecture]{
    \includegraphics[width=0.98\linewidth]{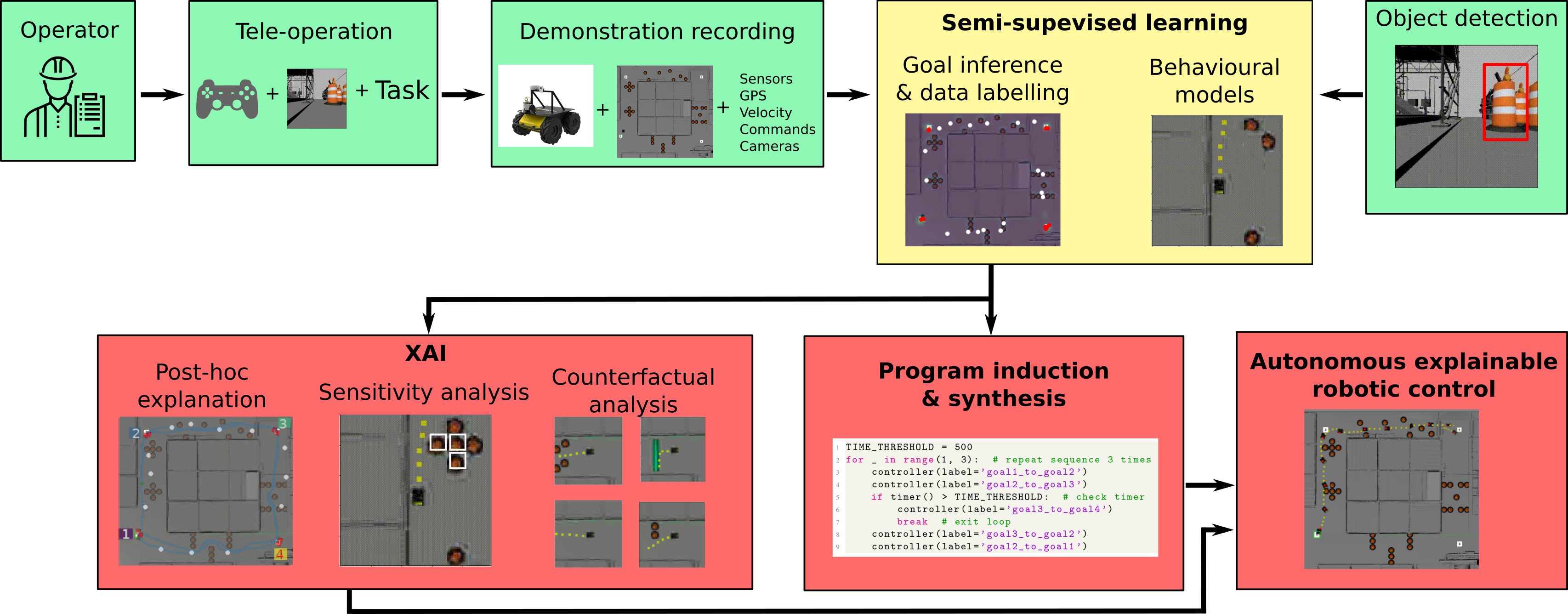}
    \label{fig:block-diagram}
  }
    \centering
    \subfigure[Husky robot]{
        \includegraphics[width=0.20\linewidth]{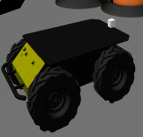}
        \label{fig:demo-husky}
    }\hfill
    \subfigure[Digital twin]{
        \includegraphics[width=0.20\linewidth]{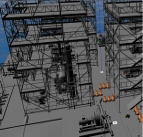}
        \label{fig:demo-dt}
    }\hfill
    \subfigure[On-board camera]{
        \includegraphics[width=0.20\linewidth]{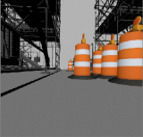}
        \label{fig:demo-onboard}
    }\hfill
    \subfigure[Demonstration]{
        \includegraphics[width=0.28\linewidth]{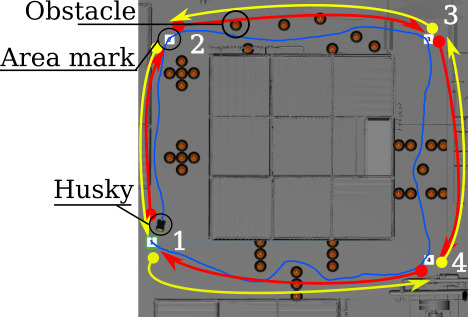}
        \label{fig:demo-demo}
    }
    \caption{Image (a) shows the architecture of the system. The green blocks represent activities to record the operator demonstrations and the object detection model that are inputs to the inference system. The yellow block represents the module of the system that trains the models, automatically labels the demonstration data and performs  goal inference. The red blocks are outputs of the system including the explainable AI module (XAI), program induction and synthesis, and finally autonomous robot control. Image (b) shows the \emph{Husky} robot in the Gazebo simulator. (c) represents the oil rig digital twin~\cite{pairet2019digital}. Image (d) shows the on-board camera used by the operator as a visual signal. (e) is the top-view of the oil rig, the operator is asked to manoeuvre the robot from one area to the next following the red arrows, starting on area 1. When the robot reaches area 1 again it reverts the sequence (yellow U-turn arrow). The blue line is the actual path that the operator followed for one demonstration (showing only the first half in the image).
    \label{fig:demo}}
\end{figure}

Fig.~\ref{fig:block-diagram} shows a diagram of the architecture of the system with all the modules and their interactions.

\section{Related Work}
Learning from demonstration (LfD) is a useful paradigm for robot programming~\cite{argall2009survey}. Initial work in this area focused on direct replication of motions. Since then, the focus has shifted to more general schemes aimed at producing more robust policies~\cite{atkeson1997robot}. These approaches include linear dynamical attractor systems~\cite{dixon2004trajectory}, dynamic motion primitives~\cite{pastor2009learning}, conditionally linear Gaussian models~\cite{levine2014learning,chiappa2010movement} and sparse online Gaussian processes~\cite{grollman2008sparse,butterfield2010learning}.

Trajectory optimisation approaches have been extended to end-to-end learning with deep neural networks~\cite{levine2016end}, resulting in robust task level visuomotor control through guided policy search. End-to-end learning has facilitated one-shot learning for domain transfer from human video demonstration~\cite{yu2018one} and the use of reinforcement learning for optimised control policies~\cite{rajeswaran2017learning,zhu2018reinforcement}.

End-to-end learning shows consistent improvement performance in several areas, but some of its drawbacks include a lack of interpretability, the significant amount of training data required~\cite{pinto2016supersizing}, and the fact that such models may not easily exhibit features we associate with higher-level conceptual learning and reasoning~\cite{konidaris2009skill, lake2015human}.

Another obstacle in end-to-end systems is a lack of flexibility. This  drawback is especially noticeable when the goal of a task is modified~\cite{lake2017building}, or when a systematic difference between test and training data is present~\cite{marcus2018algebraic}. Common  means of addressing these challenges are hybrid systems combining high-level symbolic reasoning with sub-symbolic machine learning systems~\cite{lake2017building, Burke19Explanation}. In this category of combined systems, semi-supervised learning is used to extract high-level structures in an unsupervised way and then use the extracted information to learn low-level representations in a supervised fashion~\cite{chapelle2009semi}.

Semi-supervised learning approaches include self-training methods~\cite{bauer1999empirical}, generative models~\cite{adiwardana2016using, springenberg2015unsupervised}, and graph- and vector-based methods~\cite{bagherzadeh2019review}. The most common objective in semi-supervised learning is to directly improve the performance of the supervised learning part~\cite{goldberg2006seeing, li2011learning}. The information that can be extracted in the unsupervised training process is assessed by the increase of the performance in the supervised learning counterpart~\cite{zhu2005semi}. In this paper, we use semi-supervised learning to synthesise controllers that generalise to new scenarios (Section~\ref{sec:synthesis}), and for interpretability and explainability (Section~\ref{sec:xai}).

Definitions for  explainability and interpretability still remain as an open question in the context of machine learning~\cite{lipton2018mythos}. What an explanation is, or when a system is more interpretable than another depends on aspects like the complexity of the explanation itself, the capacity of the user to understand the explanation and the role of the assessing user in the cycle of the system. The method that we propose includes several levels of explainability and interpretability based on complexity reduction, post hoc interpretation and decomposability. One way to achieve complexity reduction is to translate full traces of model predictions so it can be examined by a human in a single pass~\cite{penkov2017using,842154,zilke2016deepred}. In our method, we use this type of complexity reduction when we translate a demonstration (seen as a sequence of actions and sensory input) into a functional computer program. Most post hoc interpretations give explanations of predictions made by models in a black-box setting. Example of this type of explanation are sensitivity analysis in image classification~\cite{simonyan2013deep,wang2016dueling,ribeiro2016should}. We combine object recognition and sensitivity analysis to highlight the objects that are taken into consideration by the MPC. Decomposability explains the parts of the system (input data, calculation and parameters) in separated ways making them more intelligibles~\cite{lou2012intelligible,sato2001rule}. In the case of our system, the program induction and MPC are based on the decomposition of the demonstration as visually grounded goals. 

\section{Goal Inference and Program Induction}
\label{sec:goal-induction}
The first objective of our system is to identify high-level goals within a demonstration. In general, goals can be defined in different spaces of the system either explicitly or implicitly, and are task-dependent. For example, they could be key positions in an environment, the pose or joint configuration of a robot arm, continuous application of force, or some unknown reward function~\cite{ng2000algorithms}. In this work, we consider a task requiring a robot to inspect an industrial oil rig platform. The robot has to visit  pre-defined areas in the environment in a certain order. For a single demonstration, a user teleoperates a robot to visit these areas in the required order. The system then searches for goals within this demonstration trace, identifying visually salient dynamical attractors (or repellers) in the environment. First, using a probabilistic generative model, the system infers low-level controllers. Then, these controllers are visually grounded using a perception network to allow generalisation among different demonstrations and to distinguish between transient and attractor goals.

The scenario where we test our approach is an immersive teleoperation inspection task in a digital oil rig twin~\cite{pairet2019digital}. In this scenario, an operator drives an unmanned ground vehicle~\cite{husky} for inspection of different areas of the platform. The operator inspects four areas in a cyclic order, reversing direction when it reaches the starting area (Fig.~\ref{fig:demo-demo}). A demonstration trace includes the position of the robot in the scene, the actions are chosen by the operator (move forward, backwards, rotate to the left, rotate to the right), and images from top-view and on-board cameras recorded at 10Hz. 

\subsection{Sequential Importance Sampling with Attribution Prior}
%The first goal of our system is to find a series of controllers that satisfies the demonstration. 
To find the goals, we model the demonstration as a switching high-level task comprised of sub-tasks implemented by low-level controllers. Following~\cite{Burke19Explanation}, we assume that any task can be modelled as a set of proportional controllers. The state-space includes the horizontal and vertical dimensions on a top-view image of the full scenario (Fig.~\ref{fig:demo-demo}), and the rotation range in the horizontal plane relative to the robot.  The proportional controllers determine the linear and angular velocities of the robot:
%\textcolor{red}{esto se puede mejorar basado en RSS A. Generative task models}
\begin{equation}
  \mathbf u_{\mathbf x} = K^j_p\left(\mathbf x - \mathbf x^j_d\right),
\end{equation}
with $j = 1 \dots J$ denoting a sequence of controllers, $\mathbf x$ as the actual state, and $\mathbf x^j_d$ as the objective state and  gains $K^j_p$ of controller $j$.

As part of the controller inference, we model the influence of the objects in the demonstrations. The operator can guide the robot towards an object that marks the middle of the inspected area, and also avoid objects when traversing from one area to the next. We ground objects with a visual sensory network that takes the top-view image of the environment and outputs the predicted position of the robot depending on the action chosen by the operator. We train a deep convolutional network with data collected from several demonstrations. The training pairs include the image of the environment as input and the following 5 relative positions of the robot in intervals of 2s as the output. After training, we apply causal saliency detection to the image. Saliency detection allows the system to identify what objects are taken into account  by the network to predict future positions of the robot and separates them from other non-important objects like the background.
We use gradient-based sensitivity map to create a filter that highlights the salient input in the deep neural network position prediction task~\cite{selvaraju2017grad,zeiler2014visualizing}.
We use this filter as the attribution prior $\Phi$ for sequential importance sampling. In the prior, we include the Euclidean distance between $\mathbf{x}_d^k$ and the closest position to the trace in the horizontal plane. We include this distance to reduce the probability that the system infers an obstacle as an actual goal.

We model the switch of the actual controller $j$ to the next one $j+1$ as a Bernoulli trial with switch probability $p$~\cite{Burke19Explanation}. When a switch occurs, we sample goals and gains from the prior distribution $\Phi$. If no switch has occurred, we sample from Gaussian jitter. The generative model is:
\begin{align*}
  k &\sim \text{Bernoulli}(p) \\
  \mathbf x^j_d(t) &\sim 
  \begin{cases}
    \mathcal{N} (\mathbf x^j_d(t-1), \mathbf{Q_x}) & \text{if $k=0$} \\
    \Phi(\mathbf x) & \text{if $k=1$}
  \end{cases} \\
  K^j_p(t) &\sim \mathcal N (K^j_p(t-1), \mathbf{Q}_{kp})\\
  \mathbf {u_x} &\sim \mathcal N (K^j_p(\mathbf x - \mathbf x^j_d), \mathbf R),
\end{align*}
where for the sampling process, we define $\mathbf{Q}_{\mathbf{x}}$ and $\mathbf{Q}_{kp}$ as transition uncertainty terms and $\mathbf{R}$ as the controller noise. 

Algorithm~\ref{alg:sis} describes the sequential importance sampling re-sampling procedure used to infer the controller gains for a trace of $T$ states. Parameter $p$ models the probability of the robot switching to a new controller. For each step of the demonstration, the algorithm samples $N$ particles and evaluates the likelihood of the controller to satisfy the objective $\mathbf x_d^j$ from the actual state $\mathbf x$. This evaluation is made in two steps. First, for $pN$ particles, where $p$ represents the probability that the algorithm continues with the same controller to complete the task. The new particle is sampled from the previous time step particle adding Gaussian jitter. Second, for $(1 - p)N$ particles, the system samples new particles from the attribution prior $\Phi$. 

\begin{algorithm}
\caption{Sequential importance sampling with attribution prior\label{alg:sis}}
\begin{algorithmic}
\STATE Initialise $N$ particles
\FOR{$t=0$ \TO $T$}
\FOR{$k=0$ \TO $pN$}
\STATE Sample $\mathbf{x}^k_d(t) \sim  \mathcal{N}(\mathbf{x}^k_d(t-1),\mathbf{Q}_{\mathbf{x}})$
\STATE Sample $K^k_p(t) \sim \mathcal{N}(K^k_p(t-1),\mathbf{Q}_{kp})$
\STATE Evaluate $L^k = \mathcal{N}(K^j_p (\mathbf{x} - \mathbf{x}^j_d),\mathbf{R})$
\ENDFOR
\FOR{$k=pN$ \TO $N$}
\STATE Sample $\mathbf{x}^k_d(t) \sim \Phi(\mathbf{x})$
\STATE Sample $K^k_p(t) \sim \mathcal{N}(K^k_p(t-1),\mathbf{Q}_{kp})$
\STATE Evaluate $L^k = \mathcal{N}(K^j_p (\mathbf{x} - \mathbf{x}^j_d),\mathbf{R})$
\ENDFOR
\STATE Draw $N$ samples: $\mathbf{x}^k_d,K^k_p \sim L^k$
\ENDFOR
\end{algorithmic}
\end{algorithm}

As mentioned before, $\Phi$ is used as prior in the re-sampling process. An additional benefit of the use of the prior is to reduce the sampling space. We define the likelihood $L_a^k$ as the probability of a particle to be a salient point in a deep neural network model trained to predict the future positions of the robot from an overhead camera. We train the neural network as an end-to-end model $I(\mathbf{x}(t))$, with convolutional and fully-connected layers.
We extend the sampling process to re-sample based on an additional attribution likelihood. This new likelihood $L_b^k$ is based on the minimum inverse distance of the particle to the path of the robot, so as to favour particles that are closer to the path taken by the robot. Finally, the attribution prior likelihoods are:
\begin{eqnarray}
  L_a^k(\mathbf{x}^k_e) &=& \frac{\partial I(\mathbf{x}(t))}{\partial I(\mathbf{x}^k_{e})}, \\
  L_b^k(\mathbf x^k_d) &=& \min_{t \leq q \leq T} \parallel \mathbf x(q) - \mathbf x^k_d \parallel^{-1},
  %d(a, b) &=& \parallel a - b \parallel
\end{eqnarray} 
where $\mathbf{x}^k_e$ is the position of the particle in the input image, and $\mathbf x^k_d$ is the position of the particle in the scene. Note that for the calculation of $L^k_d$ we only use the horizontal and vertical dimensions of the vector, and do not consider the rotation of the robot.
Algorithm~\ref{alg:rejsample} shows the  re-sampling rules.

\begin{algorithm}
\caption{Extended re-sampling from attribution prior\label{alg:rejsample}}
\begin{algorithmic}
\STATE Draw $k = 1 \hdots N_p$ samples: $\mathbf{x}^k_d \sim \left [ \mathbf{x}(t), \hdots ,\mathbf{x}(t+M)\right]$\\
  \STATE Evaluate attribution likelihood: $L_a^k(\mathbf x^k_d) 
  $\\
\STATE Draw $k = 1 \hdots N_p$ samples: $\mathbf{x}^k_d \sim L_a^k$ \\
  \STATE Evaluate attribution likelihood: $L_b^k(\mathbf x^k_d)
  $\\
\STATE Draw $k = 1 \hdots N_p$ samples: $\mathbf{x}^k_d \sim L_b^k$
\end{algorithmic}
\end{algorithm}

\begin{figure}
    \centering
    \subfigure[Controllers]{
        \includegraphics[width=0.48\linewidth]{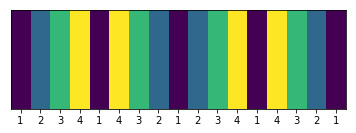}
        \label{fig:seq-1-2-c}
    }
    \subfigure[No obstacles]{
        \includegraphics[width=0.20\linewidth]{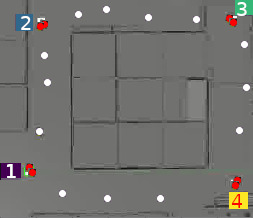}
        \label{fig:seq-1-2-a}
    }
    \subfigure[With obstacles]{
        \includegraphics[width=0.20\linewidth]{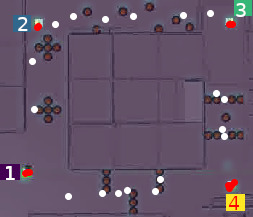}
        \label{fig:seq-1-2-b}
    }
    \caption{The colour blocks on (a) represent the sequence of inferred controllers. Each controller is associated with a cluster of visiting goals (red dots) by hand-coding each cluster with numbered colour boxes.
The white dots represent transit goals. 
    The extracted sequence of goals labelled by area is $[1, 2, 3, 4, 1, 4, 3, 2, 1]$ (repeated once) that correspond to the original demonstration. Image (b) is the scenario of a demonstration without obstacles, and (c) with them. The extracted sequence of goals is the same for visiting goals for both demonstrations. We set the constants values for the sampling process at $\mathbf{Q_x} = \mathbf Q_{kp} = \mathbf R = 0.01$ and $p=0.5$.
    \label{fig:sequence}}
\end{figure}

After inference over all the steps $T$, the result is a distribution over possible controllers for each time step. Using the effective particle size~\cite{kong1994sequential},
\begin{equation}
  N_{\text {eff}} = \frac{1}{\sum^N_{k=0}(L^k)^2}
  \label{eq:neff}
\end{equation}
we can isolate a sequence of controllers that constitute the demonstration. The idea behind effective particle size is that, on one hand, when the demonstration is switching to a new goal, the majority of particles will have a low probability mass. The low probability of the particles is represented by low effective particle size. On the other hand, when only a single controller accumulates most of the probability mass, the effective particle size will be greatest, and will have established a clear controller goal and gains. Using a peak detector over the effective particle sizes for all the time steps of the demonstration, we can identify the switching of controllers with their associated goal. We identify a sequence of controllers by maximum a-posteriori controller selection at each of the peaks. Using a K-means algorithm with elbow criterion for the number of clusters, the controllers are grouped to produce a symbolic behavioural trace of the demonstration. We asked the operator to run two full-cycles of the demonstration starting in area 1, following a clock-wise visit of the other corners, and inverting the direction once reached area 1 again (Fig.~\ref{fig:demo-demo}). Figure~\ref{fig:sequence} shows as circles the maximum a-posteriori controllers that the system identified. White circles represent low-saliency goals. We term these goals transit goals. The red circle identify high-saliency points clustered as visiting goals. The visiting goals properly delimit the primitives that define the demonstration and the visiting order. 

Once the goals are identified, we induce a program that represents an abstraction of the demonstration (Listing~\ref{lis:program1}).
To simplify the program, we search for repetition of sub-sequences (loops) and palindromes.
A robot operator (with programming skills) can use this program to examine the demonstration. 

\begin{minipage}{.96\textwidth}
  \begin{lstlisting}[language=Python, caption={Induced program from demonstration. The \emph{execute(g)} command is a call to a controller that drives the robot to the \emph{g} goal.
}, label=lis:program1]
def program():
	execute(1)
	for j in range(2):  # Demonstration loop
		controller_list = [2,3,4,1]
		count = 0
		for k in range(len(controller_list)*2-1):  # Palindromic sequence
			execute(controller_list[count])
			if k >= len(controller_list)-1:
				count = count-1
			else:
				count = count+1
		execute(1)
	return
\end{lstlisting}
\end{minipage}

\begin{figure}
    \centering
    \subfigure[Keep on the left side.]{
        \includegraphics[height=90pt]{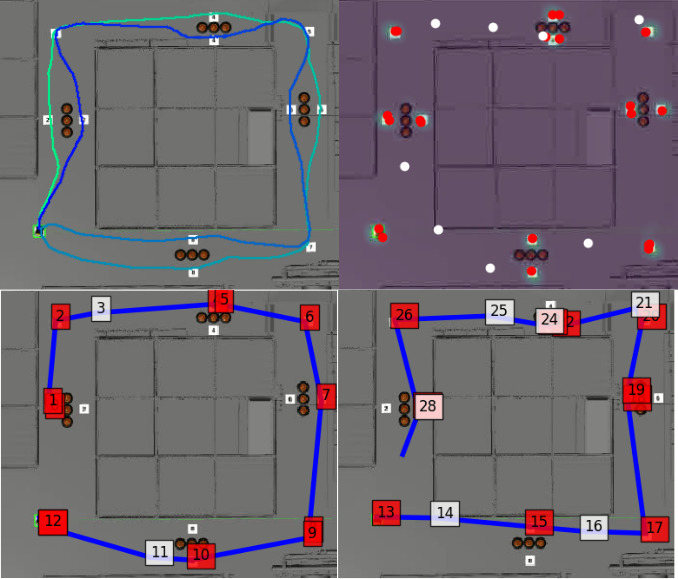}
        \label{fig:scenario4a}
    }
    \hfill
    \subfigure[Incremental navigation.]{
        \includegraphics[height=90pt]{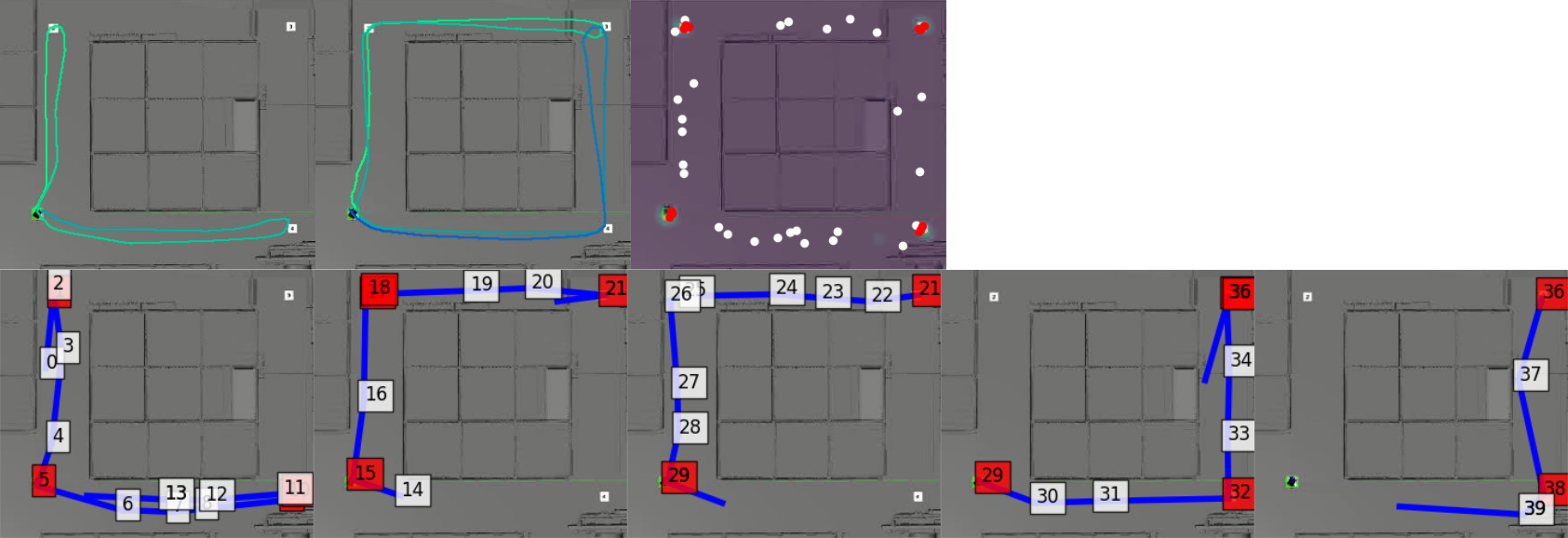}
        \label{fig:scenario4b}
    }
    \\
    \subfigure[Arrive to the other side and return.]{
        \includegraphics[height=90pt]{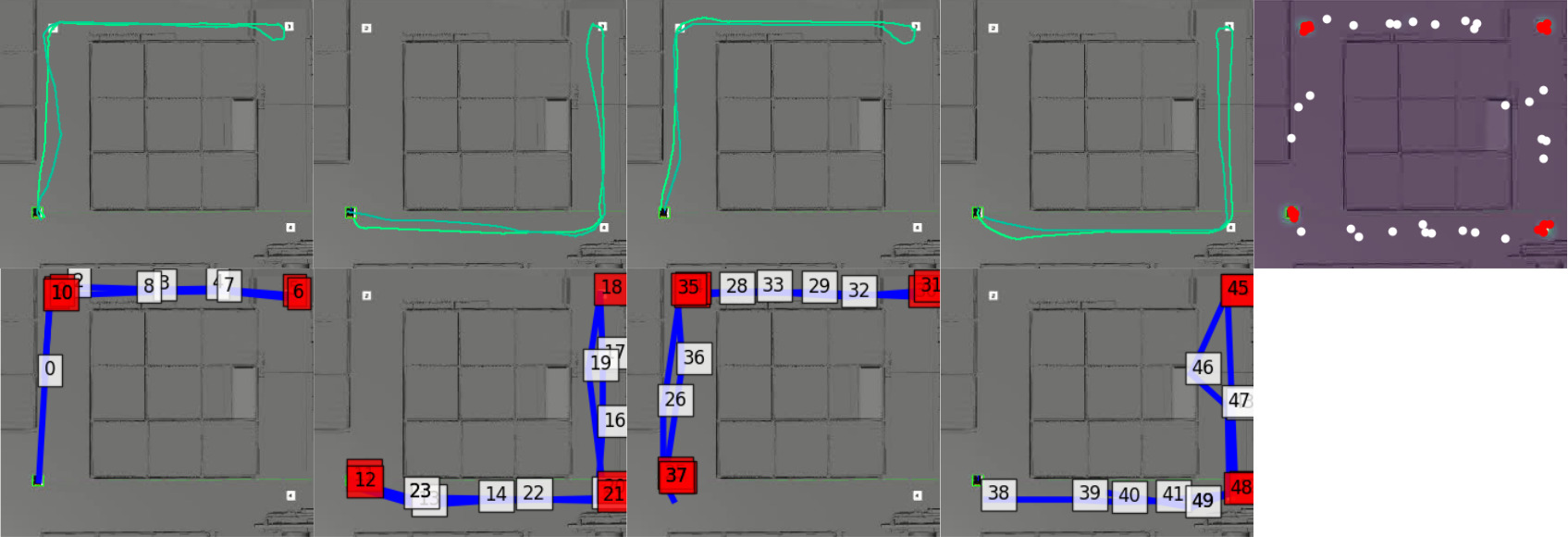}
        \label{fig:scenario4c}
    }
    \caption{
        Extended surveillance scenarios. For all of them, we show the recorded position of the tele-operated demonstration (top-left images), the inferred visiting and transit goals (red and white dots in the purple-background top-right image). At the bottom of each scenario the sequence of controllers (presented in multiple images). Inferred programs in Appendix~\ref{app:programs}.
    \label{fig:newscenarios}}
\end{figure}

Fig.~\ref{fig:newscenarios} presents results for three extended demonstrations. In the first scenario \emph{(a)}, we ask the operator to visit all the corners of the scenario while keeping the robot on the left lane. After arriving back to the first corner, turn back and return following the same instructions. For the second demonstration \emph{(b)}, the operator has to visit the adjacent corners and return, then visit the opposite corner (using the two available paths) and return. For the third demonstration \emph{(c)}, we ask the operator to visit the opposite corner, return, and visit it again using a different path. This behaviour has to be repeated at least once. Top-left image in Fig.~\ref{fig:scenario4a} shows the demonstration path (time from light- to dark-blue colour). The top-right image shows the visiting and transit goals inferred by our system (red and white dots respectively). The bottom images depict the sequence of the goals that are used to infer the program and the blue lines the linear controller that satisfy the transit from one goal to the next. The controllers are clustered and sequenced to build the final program that represents the demonstration. We include the programs for these scenarios in Appendix~\ref{app:programs}.
For scenarios \emph{(a)} (App.~\ref{lis:program4a}) the induced program can not be simplified and is presented as a list of controllers. For scenario \emph{(b)} (App.~\ref{lis:program4b}) two palindromes are found (where the robot turns back) and translated into an iterative loop in the program. For scenario \emph{(c)} (App.~\ref{lis:program4c}), the system finds the repetition of the sequence and the palindromes where the robot turns back, adding one level of nested loops. We argue that these programs require a lower extraneous cognitive load (amount of effort placed on the working memory during a task caused by the way the task is presented to the user~\cite{sweller2011cognitive,ahmad2019trust}) for an expert user to understand the demonstration. The program can be examined directly compared to waiting for the demonstration to be executed. Also, the goals are automatically detected by the system rather than by the user. Other ways of quantifying the readability a program are based on the human perception of the quality of the program itself~\cite{borstler2011quality,borstler2016beauty}. The quality of the program can be measured by its size, cyclomatic complexity, inter-procedural nesting or abtraction complexity, among others. That type of measure is out of the scope of this work as we are not focused on finding the most readable program, but a suitable translation from a demonstration to a program. In the next section, we show the usefulness of the program abstraction for synthesising new behaviour and for robotic control. 

The next section describes how to synthesise new programs and how to run them autonomously on the robot.

\section{Program Synthesis and Autonomous Control}\label{sec:synthesis}
In the previous section, we showed how we inferred goals from a single demonstration. The method also infers proportional controllers that satisfy the demonstration. For robotic control purposes, we would like to play the controllers back to the robot and have the demonstration autonomously repeated. This type of automation is a core part of learning from demonstration. One of the advantages of the program induction is that such representation of the demonstration allows implementing changes to generalise to new tasks. For example, new visiting orders of the goal areas, and the ease of adding loops and conditional control flow without losing generalisation. Another advantage is that the demonstration (autonomous behaviour in this case) remains interpretable goal-wise.

In practice, if the robot follows the proportional controllers, generalisation to new configurations of the environment is limited (although generalisation made by modifying the program remains). The visiting-goals controllers do not account for obstacles. Thus, those controllers are unable to generalise to new obstacle positions. Even if we include the transit goal controllers, the generalisation remains bound to the degree of modification of the environment. In order to address this, we use the inferred goals to label the data points of the demonstration. With the labelled data, we can train behavioural models that predict the next step position of the robot for each controller. Then, we can use MPC techniques~\cite{camacho2013model} so the robot can autonomously repeat the behaviour. This method has the advantage that visiting goals are invariant between demonstrations of the same task with different obstacles configuration. In the example of Fig.~\ref{fig:sequence}, the robot moves from the goal in area 2 to the goal in area 3. As the position of the goals is known, we can label the demonstration trace from the moment that the robot has arrived at the initial goal in area 2 until it reaches the goal in area 3. Note that training a network to predict the future position of the robot without labels, i.e. training several demonstrations with full traces, will result in the inability of the system to learn a meaningful representation when the behaviour is not distinguishable from another. For example, in two different demonstrations, the robot might traverse the same hall but taking a different route after reaching a corner. Our system would be able to separate these two behaviours with different labels. Without extra information, it not possible to separate these behaviour (same input, different output) with this type or architecture. Other approaches of unsupervised learning have been proposed to tackle this problem~\cite{murray1997multiple}, using Gaussian mixture models fit using expectation maximisation~\cite{dempster1977maximum} and variational approaches for switching state space models~\cite{ghahramani2000variational}.

\begin{figure}[!ht]
    \centering
    \includegraphics[width=0.7\linewidth]{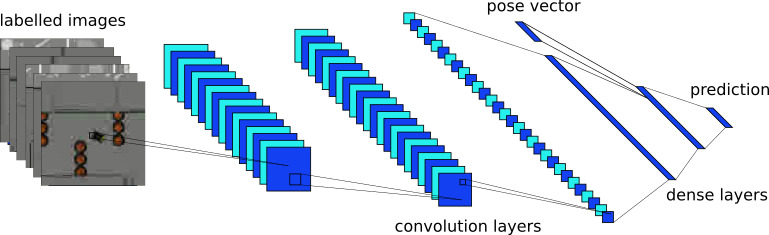}
    \caption{Diagram of the convolutional and dense layers neural network for future pose prediction. Images used as input are a close-up view of the top-view camera with the robot in the centre. The network includes three convolutional layers with a dropout rate of $0.5$. The pose information is concatenated to a dense layer output after the last convolution layer. The final output of the network is a fully-connected layer with a linear activation function. We trained a different network for each label, feeding it with data from 5 different demonstrations.
    \label{fig:nn}}
\end{figure}

We captured a set of demonstrations with the same inspection goal but with different obstacle configurations, including one scene without obstacles. Figure~\ref{fig:sequence} shows two demonstrations, one with and another without obstacles. The system infers the same visiting goals but different transit goals. We use the high-level visiting goals to label the data points within the demonstrations. To account only for obstacles close to the robot, we reduce the top-view of the scene to a $100$x$100$ pixels images centred on the robot. This reduction increases generalisation as now the behaviour is only locally based and does not depend on the configuration of the whole scene. As input to the network, we include the horizontal and vertical linear velocities and the rotation velocity over the horizontal axis of the robot. The output vector $\mathbf y_i \in \mathbb R^{10}$ includes the position of the robot after 10, 20, 30, and 50 steps (with the simulation step size of 0.1s). We trained one  network for each label, for a set of 5 demonstrations with $\sim$11,000 data pairs. Note that a high number of data points is required for the training of the deep models, but a program can be induced from a single demonstration as shown in Section~\ref{sec:goal-induction}. The network included three convolutional layers with 3 kernels each with sizes of $(7,7), (5,5)$ and $(3,3)$ with RELU activation. These layers allow the network to extract features from the images. After the feature layer we concatenate it with the pose of the robot trough a dense layer with a linear activation function~(Fig.~\ref{fig:nn}). 

Now, We can use each model to control the robot. The model prediction based controller will calculate a future position of the robot and an internal realisation of a PID controller will drive the robot to the target position. This process is repeated until the goal is reached and the next controller is invoked.

\begin{figure}
    \centering
    \includegraphics[width=0.5\linewidth]{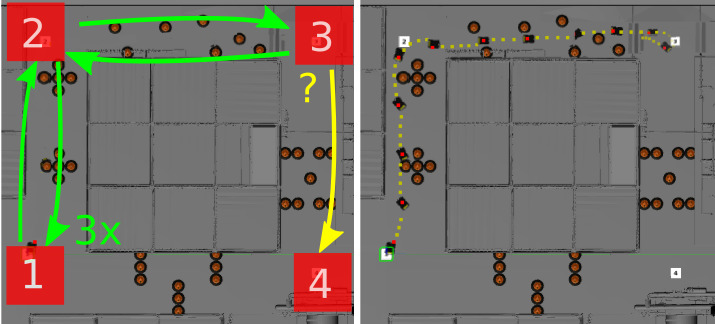}
    \caption{The image on the left depicts the synthesised program from Listing~\ref{lis:program}. The green arrows are part of the main sequence starting from $1$. The yellow arrow represents the conditional path. The red boxes are the areas where the goals are located. 
    The right image shows the synthesised program played back by the robot. The yellow dots are the predicted positions of the robots followed by the MPC. The red dots are the actual position of the robot. The image only shows the trace until line 8 of the code is reached for the first time. In this case, the conditional is not triggered. The remaining repetitions are omitted for clarity. 
    \label{fig:synthesis}}
\end{figure}

With this set of controllers (one for each model), we can synthesis new programs for behaviours not seen in the demonstration. Listing~\ref{lis:program} shows a synthesised program  (based on List.~\ref{lis:program1}) where the robot has to arrive from goal 1 to goal 3 traversing goal 2, and then reverse to return to goal 1.
This sequence can be extended with the use of conditionals and flow control. Line 2 of the program includes a loop cycle that will repeat the sequence 3 times. Line 5 includes a conditional where at arrival to goal number $3$, the system checks if enough time has passed to decide if continuing with the original sequence or call a controller that will drive the robot to goal number 4 and exit the cycle. Fig.~\ref{fig:synthesis} shows the diagram of the synthesised program and a run of the program on the robot.

%\noindent\hspace{0.15\linewidth}
\begin{minipage}{.96\textwidth}
\begin{lstlisting}[language=Python, caption={
The synthesised program, including loop and conditional control flow. The sequence of goals $(1, 2, 3, 2, 1)$ is repeated 3 times. A timer is checked if the robot has taken more than a set threshold to reach the third goal. The sequence, cycle and conditionals are not present in any demonstration, but they can be added to the synthesised program and played back by the robot. Fig.~\ref{fig:synthesis} shows one loop of the sequence without triggering the conditional.
}, label=lis:program]
TIME_THRESHOLD = 500
for _ in range(1, 3):  # repeat sequence 3 times
    controller(label='goal1_to_goal2')
    controller(label='goal2_to_goal3')
    if timer() > TIME_THRESHOLD:  # check timer
        controller(label='goal3_to_goal4')
        break  # exit loop
    controller(label='goal3_to_goal2')
    controller(label='goal2_to_goal1')
\end{lstlisting}
\end{minipage}

Even with the possibility to synthesise a more complex program, there are some restrictions to this process. For example, the use of any controller assumes that the actual position of the robot was part of the training of the model. Using a controller outside of its training domain generates undefined behaviour. For the same reason, switching from one controller to a subsequent one can only be synthesised if such switch happened in the demonstration. For example, a controller that drives the robot from goal 2 to goal 1 can only be called if the previous controller finished in goal 2. Even including these restrictions, there are enough combinations to synthesise complex behaviour from a set of few demonstrations.

We build on similar earlier work presented in~\cite{Burke19Explanation}. In that work, the execution of a same sequence or synthesised plan was restricted to scenarios where all goals were attractors. Here, we also take into account obstacle {\textit{avoidance}}, i.e., repellers. In our present approach, we have shown how the labelling of the demonstrations and the addition of repellers helps to improve generalisation. Our system is able to train a model with data from different demonstrations. This data represents the same goal (and initial condition) of a subset of the trace, but is executed in different ways and in different scenarios. The MPC scheme takes advantage of these labelling as it can execute in scenarios that are a combination of the ones seen in the demonstrations. Also, the generalisation power of convolutional neural networks improves the complexity of the scenario where the system can run autonomously.

\subsection{Learning from Demonstration Comparison}
\begin{figure}
    \centering
    \subfigure[Baseline]{
        \includegraphics[width=0.18\linewidth]{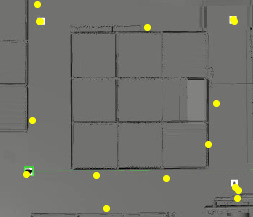}
        \label{fig:simple-scenario}
    }
    \subfigure[Baseline]{
        \includegraphics[width=0.18\linewidth]{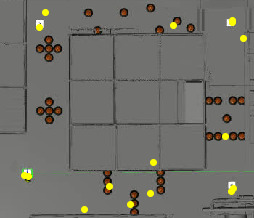}
        \label{fig:added-objects}
    }
    \subfigure[Runs]{
        \includegraphics[width=0.18\linewidth]{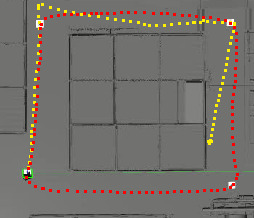}
        \label{fig:runs-2a}
    }
    \subfigure[Runs]{
        \includegraphics[width=0.18\linewidth]{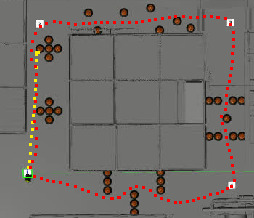}
        \label{fig:runs-2c}
    }
    \subfigure[Unseen scenario]{
        \includegraphics[width=0.18\linewidth]{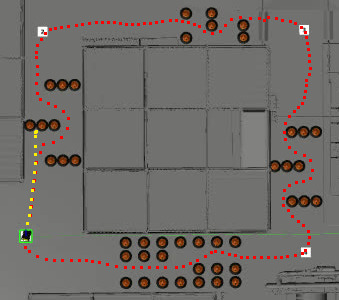}
        \label{fig:runs-3a}
    }
    \caption{Images (a) and (b) show the goals (yellow circles) inferred by the baseline. Images (c) and (d) are runs of autonomous control using the baseline proportional controllers (yellow dots) and our MPC approach (red dots). Image (e) shows the two runs on a scenario where both systems inferred the goals, but the demonstration data was not used to train the behavioural models (only the previous demonstrations). This shows the generalisation capacity of our approach. The baseline system fails to infer goals that are suitable for autonomous control. 
    \label{fig:comparison}}
\end{figure}

We compare our LfD implementation to~\cite{Burke19Explanation} as a baseline under the same scenarios. In the baseline, the authors fit a sequence of controllers using sequential importance sampling under a generative switching proportional controller task model. Compared to our system, the goal inference of the baseline does not take into account the distance to the future position of the robot or the saliency of the objects to discriminate between transition and visited goals.
Figs.~\ref{fig:simple-scenario}~and~\ref{fig:added-objects} show the goals inferred by the baseline system in the scenarios with and without obstacles. The same demonstrations have been used by our system to infer demonstration goals (see Fig.~\ref{fig:sequence}). In the scenario without obstacles (Fig.~\ref{fig:simple-scenario}), there are goals (yellow dots) outside the path of the robot and also goals over walls. The inferred goals in the scenario with obstacles (Fig.~\ref{fig:added-objects}) suffers from the same problem, i.e. goals outside a demonstrated path and over objects that have not been visited. When compared to our results, we can see that the inferred goals are closer to the demonstrated paths and are separated between visited (red dots) and transit ones (yellow dots). Our system takes into account the distance between the proposed controllers and the real path of the robot to produce more representative goals of the demonstration. Also, after inference, we include visual grounding of the goals to differentiate between goal types.

We compare the behaviour of the robot under autonomous control. For the baselines, we use the inferred controllers to autonomously drive the robot through the scenario. In our approach, we use the behavioural models learned from the semi-supervised training (Section~\ref{sec:synthesis}). Figs.~\ref{fig:runs-2a}~and~\ref{fig:runs-2c} show the runs by the baseline in yellow, and in red by our system. The baseline follows direct lines between the goals even deviating from the original demonstrations. Also, the baseline is unable to complete a full cycle as the controllers drive the robot to goals that are in the same position as obstacles or walls. To show the generalisation capacity of our approach, we run both approaches in a scenario where goals have been inferred but it has not been used to train the behavioural models. In this case, the behavioural models were trained with demonstration data from previous scenarios. Fig.~\ref{fig:runs-2c} shows that the robot can finish a cycle with our MPC approach (red dots), while the baseline is unable to react to obstacle so the robot fails to arrive at the goal destination.

\section{Explanations from Demonstration}\label{sec:xai}
In Section~\ref{sec:goal-induction}, we  showed how to extract high-level goals and induce a program from a demonstration. These results already act as an interpretation of the demonstrations.  The system abstracts dynamical attractors, modularises and translates the behavioural trace to a language understandable by an operator. An external observer can check the sequence of the demonstration by scrutinising only a few lines of codes instead of observing the full demonstration.  Another advantage is that the induced program allows operators to track the inspection process. During the running of the program or by a repetition of the demonstration, the system is capable to establish the actual stage of the inspection, indicating origin and destination goal.

\subsection{Causal Analysis}
We complement the explanations with black-box analysis~\cite{samek2017explainable} of the input in the prediction model $I(\mathbf x)$. Using causal sensitivity analysis~\cite{kim2017interpretable} and object classification, we are able to identify the objects that most influence the predictions. In this case, the predictions of future position and rotation of the robot have been learned by the model following the demonstration. For autonomous control (Section~\ref{sec:synthesis}), this analysis is a direct explanation of the behaviour of the robot. In the case of teleoperation of the robot, this analysis is an educated guess of the objects that influenced the actions chosen by the operator. This post hoc analysis is helpful when there is no access to query the operator directly, but there is access to a recording of the demonstration.

\begin{figure}
    \centering
    \subfigure[Heat map]{
        \includegraphics[width=0.18\linewidth]{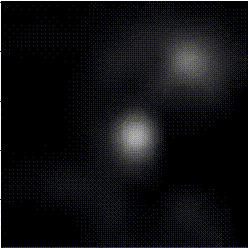}
        \label{fig:xai-heatmap}
    }
    \subfigure[Detected objects]{
        \includegraphics[width=0.18\linewidth]{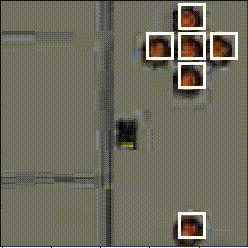}
        \label{fig:xai-do}
    }
    \subfigure[Filtered objects]{
        \includegraphics[width=0.18\linewidth]{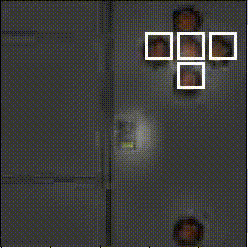}
        \label{fig:xai-fo}
    }
    \subfigure[On-board camera]{
        \includegraphics[width=0.18\linewidth]{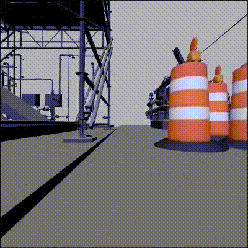}
        \label{fig:xai-obc}
    }
    \subfigure[Explanation]{
        \includegraphics[width=0.18\linewidth]{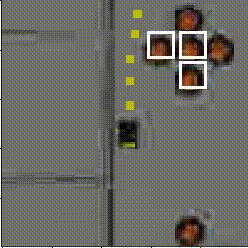}
        \label{fig:xai-exp}
    }
    \caption{
      (a) heat map of image saliency for a model prediction. (b) all objects in the vicinity of the robot. (c) objects filtered by the heat map. (d) on-board camera. (e) final result of applying the filters, including filtering objects that only appear on the on-board camera and the prediction of future position of the robot. The detection of salient objects is an explanation of the objects taken into consideration by either the operator while controlling the robot or by the models while in autonomous control.
    \label{fig:xai}}
\end{figure}

For sensitivity analysis,  we modify the input image of a single prediction with a black patch of the same size of the robot in the local top-view camera. We apply a single patch with a stride of half the size of the patch until covering the whole image. For each placement of the patch, we use the Euclidean distance between the initial prediction and the prediction made with the modified image, as the pose prediction are in Euclidean space. After applying the patch in all the positions, we build a normalised saliency heat map (Fig.~\ref{fig:xai-heatmap}). In the filter, white colour represents blobs of salient pixels. Then, we detect objects in the image (Fig.~\ref{fig:xai-do}) and filter these with a binary decision based on the saliency in the heat map. We tried several values for the threshold for the binary decision. Among those values, a threshold of $0.8$ yielded the best results, in our dataset, to include salient object-based in the size of the white blobs in the heat map (Fig.~\ref{fig:xai-fo}). A lower value for the threshold would add blobs that do not correspond to object, and a higher value would discard all objects as salient. At the next step, we apply a second filter to only represent objects appearing on the on-board camera (Figs.~\ref{fig:xai-obc}~and~\ref{fig:xai-exp}). This filter relates to the actual image that the operator perceives in an immersive demonstration. In the virtual environment of the digital twin setup, for practical reasons we assume access to the actual position of the object. However, the system can use widely available computer vision techniques for object detection like supervised learning with convolutional neural networks and 3D SLAM~\cite{krizhevsky2012imagenet, qi2017pointnet}. This is the case for real application including robots deployed in industrial environments. Our approach is modular so implementations like~\cite{liu2016ssd,ren2015faster,redmon2018yolov3} can be used when there is no direct access to the position of the objects.

The result of the causal analysis~(Fig.~\ref{fig:xai-exp}) shows both a prediction of the most likely future behaviour of the robot (yellow dots) and the objects that the controller used to decide the next action (white squares).
With this information, the user can have a better understanding of teleological and failed behaviour. This information is useful for designing new environments and new tasks that can be both automatised or human-operated. Also, these analyses can give lights to understand why an undesired action was taken by the robot and in case of failure, reduce the space of possible causes.

\subsection{Counterfactuals Explanation}
A counterfactual explanation describes the results of hypothetical cases compared to real ones. These characteristics can be a key component for reliability in an autonomous system. Counterfactuals can be used to test the limits of the system. For example, they could be used to find the smallest change to the input of a neural network that changes the classification outcome~\cite{wachter2017counterfactual}. In our system, the predictive model can be used for counterfactual causal inference. This inference can help an operator to check the consistency of the system under different scenarios before deployment. For example, in the inspection scenario, the operator can test how sensitive the system is to changes in the position of obstacles. Counterfactuals can also be used  to check how the scenario can be modified and still solve the original task. For the inspection task, if the robot fails to arrive at the desired area, the operator can modify the input to the system searching for a configuration that would have  allowed the robot to reach the goal.

\begin{figure}
    \centering
    \subfigure[Original behaviour]{
        \includegraphics[width=0.18\linewidth]{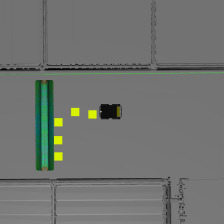}
        \label{fig:counter-ok}
    }
    \subfigure[Modifications of the original scenario by removing or adding objects.]{
      \includegraphics[width=0.18\linewidth]{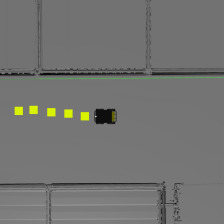}\hspace{3pt}
        \includegraphics[width=0.18\linewidth]{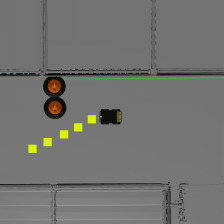}\hspace{3pt}
        \includegraphics[width=0.18\linewidth]{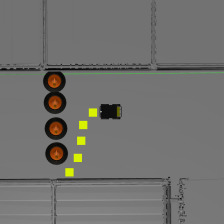}\hspace{3pt}
        \includegraphics[width=0.18\linewidth]{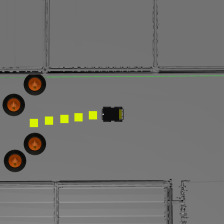}
        \label{fig:counter-exp}
    }
    \caption{
      (a) shows the prediction (yellow dots) of the positions for the robot when a green barrier is in front of it.
      (b) modifications to the original image by removing the barrier or adding new objects in different positions. The new predictions allow the operator to: i) causally infer that the barrier is responsible for the original behaviour, and ii) find a solution for the robot to continue to the goal.
    \label{fig:counter}}
\end{figure}

To test our system, we trained the behavioural model with extra demonstrations where obstacles block the passage of the robot. In these cases, the teleoperator has to find new routes or cancel the operation. In~\cite{wachter2017counterfactual,van2019interpretable}, the authors minimise a distance function between the actual state and a modified one. To find the minimum, the authors use exhaustive search over the state space, using high-level features for dimensionality reduction. 
In our approach, we ask an operator  to modify the input image at the pixel level. The modifications include adding and removing objects, and change the position of existing objects. The operator can assess each new configuration and compare the results from the model. Fig.~\ref{fig:counter-ok} shows a failed attempt to advance from left to right as a green barrier is obstructing the passage. During demonstration time, the operator consistently took a U-turn when faced with this obstacle configuration. The model learned this behaviour and correctly predicts the future position of the robot resembling the demonstrations. Now, the operator has the chance to explore a new obstacle configuration by modifying the input image. With each new configuration, the model predicts  future positions. For example, in the first modification presented on Fig.~\ref{fig:counter-exp} the barrier has been removed and the behaviour prediction indicates that the robot can continue moving to the left. The other images in Fig.~\ref{fig:counter-exp} show different obstacles placed in front of the robot with their corresponding predictions. All these predictions happen at the model level by modification of the image and do not require changes in the actual environment. This feature is useful when the modification of the environment requires a large effort, for example, in real-life scenarios where moving the obstacles would require human intervention. In these cases, more complex techniques than direct pixel modification are available, e.g. region filling or object removal~\cite{criminisi2004region, criminisi2003object}. These techniques  can be incorporated into the system to obtain the counterfactual conditionals when a top-view camera with a consistent background is not available.

\section{Conclusion}

We presented a system for program induction and explainability in learning from demonstration settings in a surveillance scenario with access to proprioceptive and onboard sensor signals. Our approach follows~\cite{Burke19Explanation}, extending it by adding dynamical repellers for obstacles, and a new controller learning process that allows us to synthesise new (obstacle aware) programs that can be successfully run by a robot without external intervention in the presented digital twin scenario.

We have presented a semi-supervised system that abstract goals and proportional controllers from an immersive teleoperated demonstration of an inspection  scenario. With this information extracted from the demonstrations, we have been able to induce a program that represents the original demonstration in a programming language for an industrial inspection task.

From  the induced program, we were able to extend the demonstrations to new tasks. We showed how new programs can be synthesised and autonomously run on the robots. In a semi-supervised approach, we use the inferred goals within the  trace to label the demonstration. This labelling allowed us to train end-to-end models that are used for model predictive control. With the synthesis of programs and the generalisation ability of the deep neural network models, we presented a subset of new programs that where applied to unseen environment. This generalisation has the potential to spawn a wide range of new behaviours and complex tasks from a reduced set of demonstrations.

Also, we presented two approaches on how to use the trained models to explain the demonstrations and autonomous behaviour. These explainable features add post hoc explanations for black-box models, complexity reduction and decomposition of the demonstration.
Through the exploitation of the low-level end-to-end models, we have been able to build a more informative system that makes explicit the salient objects and present behavioural counterfactuals.

Our approach used position on the plane as prior for goals. This approach can be generalised to other domains, e.g. 3D spaces or joint angle manipulator control. A task like tower assembly with joint angles as state-space~\cite{Burke19Explanation} can be extended to include obstacles with our approach. In general, if the goals can be satisfied by proportional controllers then this method can be applied. For more complex behaviour, techniques like Dynamic Movement Primitives~\cite{schaal2006dynamic} can be implemented.

\nocite{*}
\bibliographystyle{eptcs}
\bibliography{root}

\begin{thebibliography}{10}
\providecommand{\bibitemdeclare}[2]{}
\providecommand{\surnamestart}{}
\providecommand{\surnameend}{}
\providecommand{\urlprefix}{Available at }
\providecommand{\url}[1]{\texttt{#1}}
\providecommand{\href}[2]{\texttt{#2}}
\providecommand{\urlalt}[2]{\href{#1}{#2}}
\providecommand{\doi}[1]{doi:\urlalt{http://dx.doi.org/#1}{#1}}
\providecommand{\bibinfo}[2]{#2}

\bibitemdeclare{inproceedings}{adiwardana2016using}
\bibitem{adiwardana2016using}
\bibinfo{author}{D~\surnamestart Adiwardana\surnameend},
  \bibinfo{author}{Akihiro \surnamestart Matsukawa\surnameend} \&
  \bibinfo{author}{Jay \surnamestart Whang\surnameend} (\bibinfo{year}{2016}):
  \emph{\bibinfo{title}{Using generative models for semi-supervised learning}}.
\newblock In: {\sl \bibinfo{booktitle}{Medical image computing and
  computer-assisted intervention--MICCAI}}, \bibinfo{volume}{2016}, pp.
  \bibinfo{pages}{106--14}.

\bibitemdeclare{article}{ahmad2019trust}
\bibitem{ahmad2019trust}
\bibinfo{author}{Muneeb~Imtiaz \surnamestart Ahmad\surnameend},
  \bibinfo{author}{Jasmin \surnamestart Bernotat\surnameend},
  \bibinfo{author}{Katrin \surnamestart Lohan\surnameend} \&
  \bibinfo{author}{Friederike \surnamestart Eyssel\surnameend}
  (\bibinfo{year}{2019}): \emph{\bibinfo{title}{Trust and Cognitive Load During
  Human-Robot Interaction}}.
\newblock {\sl \bibinfo{journal}{arXiv preprint arXiv:1909.05160}}.

\bibitemdeclare{article}{argall2009survey}
\bibitem{argall2009survey}
\bibinfo{author}{Brenna~D \surnamestart Argall\surnameend},
  \bibinfo{author}{Sonia \surnamestart Chernova\surnameend},
  \bibinfo{author}{Manuela \surnamestart Veloso\surnameend} \&
  \bibinfo{author}{Brett \surnamestart Browning\surnameend}
  (\bibinfo{year}{2009}): \emph{\bibinfo{title}{A survey of robot learning from
  demonstration}}.
\newblock {\sl \bibinfo{journal}{Robotics and autonomous systems}}
  \bibinfo{volume}{57}(\bibinfo{number}{5}), pp. \bibinfo{pages}{469--483},
  \doi{10.1016/j.robot.2008.10.024}.

\bibitemdeclare{inproceedings}{atkeson1997robot}
\bibitem{atkeson1997robot}
\bibinfo{author}{Christopher~G \surnamestart Atkeson\surnameend} \&
  \bibinfo{author}{Stefan \surnamestart Schaal\surnameend}
  (\bibinfo{year}{1997}): \emph{\bibinfo{title}{Robot learning from
  demonstration}}.
\newblock In: {\sl \bibinfo{booktitle}{ICML}}, \bibinfo{volume}{97},
  \bibinfo{organization}{Citeseer}, pp. \bibinfo{pages}{12--20}.

\bibitemdeclare{article}{bagherzadeh2019review}
\bibitem{bagherzadeh2019review}
\bibinfo{author}{Jamshid \surnamestart Bagherzadeh\surnameend} \&
  \bibinfo{author}{Hasan \surnamestart Asil\surnameend} (\bibinfo{year}{2019}):
  \emph{\bibinfo{title}{A review of various semi-supervised learning models
  with a deep learning and memory approach}}.
\newblock {\sl \bibinfo{journal}{Iran Journal of Computer Science}}
  \bibinfo{volume}{2}(\bibinfo{number}{2}), pp. \bibinfo{pages}{65--80},
  \doi{10.1007/s42044-018-00027-6}.

\bibitemdeclare{article}{bauer1999empirical}
\bibitem{bauer1999empirical}
\bibinfo{author}{Eric \surnamestart Bauer\surnameend} \& \bibinfo{author}{Ron
  \surnamestart Kohavi\surnameend} (\bibinfo{year}{1999}):
  \emph{\bibinfo{title}{An empirical comparison of voting classification
  algorithms: Bagging, boosting, and variants}}.
\newblock {\sl \bibinfo{journal}{Machine learning}}
  \bibinfo{volume}{36}(\bibinfo{number}{1-2}), pp. \bibinfo{pages}{105--139},
  \doi{10.1023/A:1007515423169}.

\bibitemdeclare{article}{Billard:2013}
\bibitem{Billard:2013}
\bibinfo{author}{A.~\surnamestart Billard\surnameend} \&
  \bibinfo{author}{D.~\surnamestart Grollman\surnameend}
  (\bibinfo{year}{2013}): \emph{\bibinfo{title}{{R}obot learning by
  demonstration}}.
\newblock {\sl \bibinfo{journal}{Scholarpedia}}
  \bibinfo{volume}{8}(\bibinfo{number}{12}), p. \bibinfo{pages}{3824},
  \doi{10.4249/scholarpedia.3824}.
\newblock \bibinfo{note}{Revision \#138061}.

\bibitemdeclare{article}{bojarski2016end}
\bibitem{bojarski2016end}
\bibinfo{author}{Mariusz \surnamestart Bojarski\surnameend},
  \bibinfo{author}{Davide \surnamestart Del~Testa\surnameend},
  \bibinfo{author}{Daniel \surnamestart Dworakowski\surnameend},
  \bibinfo{author}{Bernhard \surnamestart Firner\surnameend},
  \bibinfo{author}{Beat \surnamestart Flepp\surnameend},
  \bibinfo{author}{Prasoon \surnamestart Goyal\surnameend},
  \bibinfo{author}{Lawrence~D \surnamestart Jackel\surnameend},
  \bibinfo{author}{Mathew \surnamestart Monfort\surnameend},
  \bibinfo{author}{Urs \surnamestart Muller\surnameend},
  \bibinfo{author}{Jiakai \surnamestart Zhang\surnameend} et~al.
  (\bibinfo{year}{2016}): \emph{\bibinfo{title}{End to end learning for
  self-driving cars}}.
\newblock {\sl \bibinfo{journal}{arXiv preprint arXiv:1604.07316}}.

\bibitemdeclare{article}{borstler2016beauty}
\bibitem{borstler2016beauty}
\bibinfo{author}{J{\"u}rgen \surnamestart B{\"o}rstler\surnameend},
  \bibinfo{author}{Michael~E \surnamestart Caspersen\surnameend} \&
  \bibinfo{author}{Marie \surnamestart Nordstr{\"o}m\surnameend}
  (\bibinfo{year}{2016}): \emph{\bibinfo{title}{Beauty and the Beast: on the
  readability of object-oriented example programs}}.
\newblock {\sl \bibinfo{journal}{Software quality journal}}
  \bibinfo{volume}{24}(\bibinfo{number}{2}), pp. \bibinfo{pages}{231--246},
  \doi{10.1007/s11219-015-9267-5}.

\bibitemdeclare{article}{borstler2011quality}
\bibitem{borstler2011quality}
\bibinfo{author}{J{\"u}rgen \surnamestart B{\"o}rstler\surnameend},
  \bibinfo{author}{Marie \surnamestart Nordstr{\"o}m\surnameend} \&
  \bibinfo{author}{James~H \surnamestart Paterson\surnameend}
  (\bibinfo{year}{2011}): \emph{\bibinfo{title}{On the quality of examples in
  introductory Java textbooks}}.
\newblock {\sl \bibinfo{journal}{ACM Transactions on Computing Education
  (TOCE)}} \bibinfo{volume}{11}(\bibinfo{number}{1}), pp.
  \bibinfo{pages}{1--21}, \doi{10.1145/1921607.1921610}.

\bibitemdeclare{article}{Burke19Explanation}
\bibitem{Burke19Explanation}
\bibinfo{author}{Michael \surnamestart Burke\surnameend},
  \bibinfo{author}{Svetlin \surnamestart Penkov\surnameend} \&
  \bibinfo{author}{Subramanian \surnamestart Ramamoorthy\surnameend}
  (\bibinfo{year}{2019}): \emph{\bibinfo{title}{From Explanation to Synthesis:
  Compositional Program Induction for Learning From Demonstration}}.
\newblock {\sl \bibinfo{journal}{Robotics: Science and Systems (R:SS)}},
  \doi{10.15607/RSS.2019.XV.015}.

\bibitemdeclare{inproceedings}{butterfield2010learning}
\bibitem{butterfield2010learning}
\bibinfo{author}{Jesse \surnamestart Butterfield\surnameend},
  \bibinfo{author}{Sarah \surnamestart Osentoski\surnameend},
  \bibinfo{author}{Graylin \surnamestart Jay\surnameend} \&
  \bibinfo{author}{Odest~Chadwicke \surnamestart Jenkins\surnameend}
  (\bibinfo{year}{2010}): \emph{\bibinfo{title}{Learning from demonstration
  using a multi-valued function regressor for time-series data}}.
\newblock In: {\sl \bibinfo{booktitle}{2010 10th IEEE-RAS International
  Conference on Humanoid Robots}}, \bibinfo{organization}{IEEE}, pp.
  \bibinfo{pages}{328--333}, \doi{10.1109/ICHR.2010.5686284}.

\bibitemdeclare{book}{camacho2013model}
\bibitem{camacho2013model}
\bibinfo{author}{Eduardo~F \surnamestart Camacho\surnameend} \&
  \bibinfo{author}{Carlos~Bordons \surnamestart Alba\surnameend}
  (\bibinfo{year}{2013}): \emph{\bibinfo{title}{Model predictive control}}.
\newblock \bibinfo{publisher}{Springer Science \& Business Media},
  \doi{10.1007/978-1-4471-3398-8}.

\bibitemdeclare{article}{chapelle2009semi}
\bibitem{chapelle2009semi}
\bibinfo{author}{Olivier \surnamestart Chapelle\surnameend},
  \bibinfo{author}{Bernhard \surnamestart Scholkopf\surnameend} \&
  \bibinfo{author}{Alexander \surnamestart Zien\surnameend}
  (\bibinfo{year}{2009}): \emph{\bibinfo{title}{Semi-supervised learning
  (chapelle, o. et al., eds.; 2006)[book reviews]}}.
\newblock {\sl \bibinfo{journal}{IEEE Transactions on Neural Networks}}
  \bibinfo{volume}{20}(\bibinfo{number}{3}), pp. \bibinfo{pages}{542--542},
  \doi{10.1109/TNN.2009.2015974}.

\bibitemdeclare{inproceedings}{chiappa2010movement}
\bibitem{chiappa2010movement}
\bibinfo{author}{Silvia \surnamestart Chiappa\surnameend} \&
  \bibinfo{author}{Jan~R \surnamestart Peters\surnameend}
  (\bibinfo{year}{2010}): \emph{\bibinfo{title}{Movement extraction by
  detecting dynamics switches and repetitions}}.
\newblock In: {\sl \bibinfo{booktitle}{Advances in neural information
  processing systems}}, pp. \bibinfo{pages}{388--396}.

\bibitemdeclare{inproceedings}{criminisi2003object}
\bibitem{criminisi2003object}
\bibinfo{author}{Antonio \surnamestart Criminisi\surnameend},
  \bibinfo{author}{Patrick \surnamestart Perez\surnameend} \&
  \bibinfo{author}{Kentaro \surnamestart Toyama\surnameend}
  (\bibinfo{year}{2003}): \emph{\bibinfo{title}{Object removal by
  exemplar-based inpainting}}.
\newblock In: {\sl \bibinfo{booktitle}{2003 IEEE Computer Society Conference on
  Computer Vision and Pattern Recognition, 2003. Proceedings.}},
  \bibinfo{volume}{2}, \bibinfo{organization}{IEEE}, pp.
  \bibinfo{pages}{II--II}, \doi{10.1109/CVPR.2003.1211538}.

\bibitemdeclare{article}{criminisi2004region}
\bibitem{criminisi2004region}
\bibinfo{author}{Antonio \surnamestart Criminisi\surnameend},
  \bibinfo{author}{Patrick \surnamestart P{\'e}rez\surnameend} \&
  \bibinfo{author}{Kentaro \surnamestart Toyama\surnameend}
  (\bibinfo{year}{2004}): \emph{\bibinfo{title}{Region filling and object
  removal by exemplar-based image inpainting}}.
\newblock {\sl \bibinfo{journal}{IEEE Transactions on image processing}}
  \bibinfo{volume}{13}(\bibinfo{number}{9}), pp. \bibinfo{pages}{1200--1212},
  \doi{10.1109/TIP.2004.833105}.

\bibitemdeclare{inproceedings}{6386047}
\bibitem{6386047}
\bibinfo{author}{C.~\surnamestart {Daniel}\surnameend},
  \bibinfo{author}{G.~\surnamestart {Neumann}\surnameend} \&
  \bibinfo{author}{J.~\surnamestart {Peters}\surnameend}
  (\bibinfo{year}{2012}): \emph{\bibinfo{title}{Learning concurrent motor
  skills in versatile solution spaces}}.
\newblock In: {\sl \bibinfo{booktitle}{2012 IEEE/RSJ International Conference
  on Intelligent Robots and Systems}}, pp. \bibinfo{pages}{3591--3597},
  \doi{10.1109/IROS.2012.6386047}.

\bibitemdeclare{article}{dempster1977maximum}
\bibitem{dempster1977maximum}
\bibinfo{author}{Arthur~P \surnamestart Dempster\surnameend},
  \bibinfo{author}{Nan~M \surnamestart Laird\surnameend} \&
  \bibinfo{author}{Donald~B \surnamestart Rubin\surnameend}
  (\bibinfo{year}{1977}): \emph{\bibinfo{title}{Maximum likelihood from
  incomplete data via the EM algorithm}}.
\newblock {\sl \bibinfo{journal}{Journal of the Royal Statistical Society:
  Series B (Methodological)}} \bibinfo{volume}{39}(\bibinfo{number}{1}), pp.
  \bibinfo{pages}{1--22}.

\bibitemdeclare{inproceedings}{dixon2004trajectory}
\bibitem{dixon2004trajectory}
\bibinfo{author}{Kevin~R \surnamestart Dixon\surnameend} \&
  \bibinfo{author}{Pradeep~K \surnamestart Khosla\surnameend}
  (\bibinfo{year}{2004}): \emph{\bibinfo{title}{Trajectory representation using
  sequenced linear dynamical systems}}.
\newblock In: {\sl \bibinfo{booktitle}{IEEE International Conference on
  Robotics and Automation, 2004. Proceedings. ICRA'04. 2004}},
  \bibinfo{volume}{4}, \bibinfo{organization}{IEEE}, pp.
  \bibinfo{pages}{3925--3930}, \doi{10.1109/ROBOT.2004.1308881}.

\bibitemdeclare{inproceedings}{dovsilovic2018explainable}
\bibitem{dovsilovic2018explainable}
\bibinfo{author}{Filip~Karlo \surnamestart Do{\v{s}}ilovi{\'c}\surnameend},
  \bibinfo{author}{Mario \surnamestart Br{\v{c}}i{\'c}\surnameend} \&
  \bibinfo{author}{Nikica \surnamestart Hlupi{\'c}\surnameend}
  (\bibinfo{year}{2018}): \emph{\bibinfo{title}{Explainable artificial
  intelligence: A survey}}.
\newblock In: {\sl \bibinfo{booktitle}{2018 41st International convention on
  information and communication technology, electronics and microelectronics
  (MIPRO)}}, \bibinfo{organization}{IEEE}, pp. \bibinfo{pages}{0210--0215},
  \doi{10.23919/MIPRO.2018.8400040}.

\bibitemdeclare{article}{fu2019neural}
\bibitem{fu2019neural}
\bibinfo{author}{Yiwei \surnamestart Fu\surnameend}, \bibinfo{author}{Devesh~K
  \surnamestart Jha\surnameend}, \bibinfo{author}{Zeyu \surnamestart
  Zhang\surnameend}, \bibinfo{author}{Zhenyuan \surnamestart Yuan\surnameend}
  \& \bibinfo{author}{Asok \surnamestart Ray\surnameend}
  (\bibinfo{year}{2019}): \emph{\bibinfo{title}{Neural Network-Based Learning
  from Demonstration of an Autonomous Ground Robot}}.
\newblock {\sl \bibinfo{journal}{Machines}}
  \bibinfo{volume}{7}(\bibinfo{number}{2}), p.~\bibinfo{pages}{24},
  \doi{10.3390/machines7020024}.

\bibitemdeclare{article}{ghahramani2000variational}
\bibitem{ghahramani2000variational}
\bibinfo{author}{Zoubin \surnamestart Ghahramani\surnameend} \&
  \bibinfo{author}{Geoffrey~E \surnamestart Hinton\surnameend}
  (\bibinfo{year}{2000}): \emph{\bibinfo{title}{Variational learning for
  switching state-space models}}.
\newblock {\sl \bibinfo{journal}{Neural computation}}
  \bibinfo{volume}{12}(\bibinfo{number}{4}), pp. \bibinfo{pages}{831--864},
  \doi{10.1162/089976600300015619}.

\bibitemdeclare{inproceedings}{goldberg2006seeing}
\bibitem{goldberg2006seeing}
\bibinfo{author}{Andrew~B \surnamestart Goldberg\surnameend} \&
  \bibinfo{author}{Xiaojin \surnamestart Zhu\surnameend}
  (\bibinfo{year}{2006}): \emph{\bibinfo{title}{Seeing stars when there aren't
  many stars: graph-based semi-supervised learning for sentiment
  categorization}}.
\newblock In: {\sl \bibinfo{booktitle}{Proceedings of the first workshop on
  graph based methods for natural language processing}},
  \bibinfo{organization}{Association for Computational Linguistics}, pp.
  \bibinfo{pages}{45--52}, \doi{10.3115/1654758.1654769}.

\bibitemdeclare{article}{Gribovskaya:148817}
\bibitem{Gribovskaya:148817}
\bibinfo{author}{Elena \surnamestart Gribovskaya\surnameend},
  \bibinfo{author}{S.~M. \surnamestart Khansari-Zadeh\surnameend} \&
  \bibinfo{author}{Aude \surnamestart Billard\surnameend}
  (\bibinfo{year}{2011}): \emph{\bibinfo{title}{Learning Nonlinear Multivariate
  Dynamics of Motion in Robotic Manipulators [accepted]}}.
\newblock {\sl \bibinfo{journal}{International Journal of Robotics Research}}
  \bibinfo{volume}{30}(\bibinfo{number}{8}), pp. \bibinfo{pages}{80--117},
  \doi{10.1177/0278364910376251}.
\newblock \urlprefix\url{http://infoscience.epfl.ch/record/148817}.

\bibitemdeclare{inproceedings}{5650500}
\bibitem{5650500}
\bibinfo{author}{D.~H. \surnamestart {Grollman}\surnameend} \&
  \bibinfo{author}{O.~C. \surnamestart {Jenkins}\surnameend}
  (\bibinfo{year}{2010}): \emph{\bibinfo{title}{Incremental learning of
  subtasks from unsegmented demonstration}}.
\newblock In: {\sl \bibinfo{booktitle}{2010 IEEE/RSJ International Conference
  on Intelligent Robots and Systems}}, pp. \bibinfo{pages}{261--266},
  \doi{10.1109/IROS.2010.5650500}.

\bibitemdeclare{inproceedings}{grollman2008sparse}
\bibitem{grollman2008sparse}
\bibinfo{author}{Daniel~H \surnamestart Grollman\surnameend} \&
  \bibinfo{author}{Odest~Chadwicke \surnamestart Jenkins\surnameend}
  (\bibinfo{year}{2008}): \emph{\bibinfo{title}{Sparse incremental learning for
  interactive robot control policy estimation}}.
\newblock In: {\sl \bibinfo{booktitle}{2008 IEEE International Conference on
  Robotics and Automation}}, \bibinfo{organization}{IEEE}, pp.
  \bibinfo{pages}{3315--3320}, \doi{10.1109/ROBOT.2008.4543716}.

\bibitemdeclare{article}{hastie2018orca}
\bibitem{hastie2018orca}
\bibinfo{author}{Helen \surnamestart Hastie\surnameend},
  \bibinfo{author}{Katrin \surnamestart Lohan\surnameend},
  \bibinfo{author}{Mike \surnamestart Chantler\surnameend},
  \bibinfo{author}{David~A \surnamestart Robb\surnameend},
  \bibinfo{author}{Subramanian \surnamestart Ramamoorthy\surnameend},
  \bibinfo{author}{Ron \surnamestart Petrick\surnameend},
  \bibinfo{author}{Sethu \surnamestart Vijayakumar\surnameend} \&
  \bibinfo{author}{David \surnamestart Lane\surnameend} (\bibinfo{year}{2018}):
  \emph{\bibinfo{title}{The ORCA hub: Explainable offshore robotics through
  intelligent interfaces}}.
\newblock {\sl \bibinfo{journal}{arXiv preprint arXiv:1803.02100}}.

\bibitemdeclare{article}{hutter2017anymal}
\bibitem{hutter2017anymal}
\bibinfo{author}{Marco \surnamestart Hutter\surnameend},
  \bibinfo{author}{Christian \surnamestart Gehring\surnameend},
  \bibinfo{author}{Andreas \surnamestart Lauber\surnameend},
  \bibinfo{author}{Fabian \surnamestart Gunther\surnameend},
  \bibinfo{author}{Carmine~Dario \surnamestart Bellicoso\surnameend},
  \bibinfo{author}{Vassilios \surnamestart Tsounis\surnameend},
  \bibinfo{author}{P{\'e}ter \surnamestart Fankhauser\surnameend},
  \bibinfo{author}{Remo \surnamestart Diethelm\surnameend},
  \bibinfo{author}{Samuel \surnamestart Bachmann\surnameend},
  \bibinfo{author}{Michael \surnamestart Bl{\"o}sch\surnameend} et~al.
  (\bibinfo{year}{2017}): \emph{\bibinfo{title}{ANYmal-toward legged robots for
  harsh environments}}.
\newblock {\sl \bibinfo{journal}{Advanced Robotics}}
  \bibinfo{volume}{31}(\bibinfo{number}{17}), pp. \bibinfo{pages}{918--931},
  \doi{10.1080/01691864.2017.1378591}.

\bibitemdeclare{inproceedings}{kim2017interpretable}
\bibitem{kim2017interpretable}
\bibinfo{author}{Jinkyu \surnamestart Kim\surnameend} \& \bibinfo{author}{John
  \surnamestart Canny\surnameend} (\bibinfo{year}{2017}):
  \emph{\bibinfo{title}{Interpretable learning for self-driving cars by
  visualizing causal attention}}.
\newblock In: {\sl \bibinfo{booktitle}{Proceedings of the IEEE international
  conference on computer vision}}, pp. \bibinfo{pages}{2942--2950},
  \doi{10.1109/ICCV.2017.320}.

\bibitemdeclare{article}{kong1994sequential}
\bibitem{kong1994sequential}
\bibinfo{author}{Augustine \surnamestart Kong\surnameend},
  \bibinfo{author}{Jun~S \surnamestart Liu\surnameend} \&
  \bibinfo{author}{Wing~Hung \surnamestart Wong\surnameend}
  (\bibinfo{year}{1994}): \emph{\bibinfo{title}{Sequential imputations and
  Bayesian missing data problems}}.
\newblock {\sl \bibinfo{journal}{Journal of the American statistical
  association}} \bibinfo{volume}{89}(\bibinfo{number}{425}), pp.
  \bibinfo{pages}{278--288}, \doi{10.1080/01621459.1994.10476469}.

\bibitemdeclare{inproceedings}{konidaris2009skill}
\bibitem{konidaris2009skill}
\bibinfo{author}{George \surnamestart Konidaris\surnameend} \&
  \bibinfo{author}{Andrew~G \surnamestart Barto\surnameend}
  (\bibinfo{year}{2009}): \emph{\bibinfo{title}{Skill discovery in continuous
  reinforcement learning domains using skill chaining}}.
\newblock In: {\sl \bibinfo{booktitle}{Advances in neural information
  processing systems}}, pp. \bibinfo{pages}{1015--1023}.

\bibitemdeclare{inproceedings}{krizhevsky2012imagenet}
\bibitem{krizhevsky2012imagenet}
\bibinfo{author}{Alex \surnamestart Krizhevsky\surnameend},
  \bibinfo{author}{Ilya \surnamestart Sutskever\surnameend} \&
  \bibinfo{author}{Geoffrey~E \surnamestart Hinton\surnameend}
  (\bibinfo{year}{2012}): \emph{\bibinfo{title}{Imagenet classification with
  deep convolutional neural networks}}.
\newblock In: {\sl \bibinfo{booktitle}{Advances in neural information
  processing systems}}, pp. \bibinfo{pages}{1097--1105}, \doi{10.1145/3065386}.

\bibitemdeclare{article}{Kulic:2012:ILF:2159336.2159342}
\bibitem{Kulic:2012:ILF:2159336.2159342}
\bibinfo{author}{Dana \surnamestart Kuli\'{c}\surnameend},
  \bibinfo{author}{Christian \surnamestart Ott\surnameend},
  \bibinfo{author}{Dongheui \surnamestart Lee\surnameend},
  \bibinfo{author}{Junichi \surnamestart Ishikawa\surnameend} \&
  \bibinfo{author}{Yoshihiko \surnamestart Nakamura\surnameend}
  (\bibinfo{year}{2012}): \emph{\bibinfo{title}{Incremental Learning of Full
  Body Motion Primitives and Their Sequencing Through Human Motion
  Observation}}.
\newblock {\sl \bibinfo{journal}{Int. J. Rob. Res.}}
  \bibinfo{volume}{31}(\bibinfo{number}{3}), pp. \bibinfo{pages}{330--345},
  \doi{10.1177/0278364911426178}.

\bibitemdeclare{article}{lake2015human}
\bibitem{lake2015human}
\bibinfo{author}{Brenden~M \surnamestart Lake\surnameend},
  \bibinfo{author}{Ruslan \surnamestart Salakhutdinov\surnameend} \&
  \bibinfo{author}{Joshua~B \surnamestart Tenenbaum\surnameend}
  (\bibinfo{year}{2015}): \emph{\bibinfo{title}{Human-level concept learning
  through probabilistic program induction}}.
\newblock {\sl \bibinfo{journal}{Science}}
  \bibinfo{volume}{350}(\bibinfo{number}{6266}), pp.
  \bibinfo{pages}{1332--1338}, \doi{10.1126/science.aab3050}.

\bibitemdeclare{article}{lake2017building}
\bibitem{lake2017building}
\bibinfo{author}{Brenden~M \surnamestart Lake\surnameend},
  \bibinfo{author}{Tomer~D \surnamestart Ullman\surnameend},
  \bibinfo{author}{Joshua~B \surnamestart Tenenbaum\surnameend} \&
  \bibinfo{author}{Samuel~J \surnamestart Gershman\surnameend}
  (\bibinfo{year}{2017}): \emph{\bibinfo{title}{Building machines that learn
  and think like people}}.
\newblock {\sl \bibinfo{journal}{Behavioral and brain sciences}}
  \bibinfo{volume}{40}, \doi{10.1017/S0140525X16001837}.

\bibitemdeclare{inproceedings}{levine2014learning}
\bibitem{levine2014learning}
\bibinfo{author}{Sergey \surnamestart Levine\surnameend} \&
  \bibinfo{author}{Pieter \surnamestart Abbeel\surnameend}
  (\bibinfo{year}{2014}): \emph{\bibinfo{title}{Learning neural network
  policies with guided policy search under unknown dynamics}}.
\newblock In: {\sl \bibinfo{booktitle}{Advances in Neural Information
  Processing Systems}}, pp. \bibinfo{pages}{1071--1079}.

\bibitemdeclare{article}{levine2016end}
\bibitem{levine2016end}
\bibinfo{author}{Sergey \surnamestart Levine\surnameend},
  \bibinfo{author}{Chelsea \surnamestart Finn\surnameend},
  \bibinfo{author}{Trevor \surnamestart Darrell\surnameend} \&
  \bibinfo{author}{Pieter \surnamestart Abbeel\surnameend}
  (\bibinfo{year}{2016}): \emph{\bibinfo{title}{End-to-end training of deep
  visuomotor policies}}.
\newblock {\sl \bibinfo{journal}{The Journal of Machine Learning Research}}
  \bibinfo{volume}{17}(\bibinfo{number}{1}), pp. \bibinfo{pages}{1334--1373}.

\bibitemdeclare{inproceedings}{li2011learning}
\bibitem{li2011learning}
\bibinfo{author}{Fangtao~Huang \surnamestart Li\surnameend},
  \bibinfo{author}{Minlie \surnamestart Huang\surnameend},
  \bibinfo{author}{Yi~\surnamestart Yang\surnameend} \&
  \bibinfo{author}{Xiaoyan \surnamestart Zhu\surnameend}
  (\bibinfo{year}{2011}): \emph{\bibinfo{title}{Learning to identify review
  spam}}.
\newblock In: {\sl \bibinfo{booktitle}{Twenty-second international joint
  conference on artificial intelligence}}.

\bibitemdeclare{article}{lipton2018mythos}
\bibitem{lipton2018mythos}
\bibinfo{author}{Zachary~C \surnamestart Lipton\surnameend}
  (\bibinfo{year}{2018}): \emph{\bibinfo{title}{The mythos of model
  interpretability}}.
\newblock {\sl \bibinfo{journal}{Queue}}
  \bibinfo{volume}{16}(\bibinfo{number}{3}), pp. \bibinfo{pages}{31--57},
  \doi{10.1145/3233231}.

\bibitemdeclare{inproceedings}{liu2016ssd}
\bibitem{liu2016ssd}
\bibinfo{author}{Wei \surnamestart Liu\surnameend}, \bibinfo{author}{Dragomir
  \surnamestart Anguelov\surnameend}, \bibinfo{author}{Dumitru \surnamestart
  Erhan\surnameend}, \bibinfo{author}{Christian \surnamestart
  Szegedy\surnameend}, \bibinfo{author}{Scott \surnamestart Reed\surnameend},
  \bibinfo{author}{Cheng-Yang \surnamestart Fu\surnameend} \&
  \bibinfo{author}{Alexander~C \surnamestart Berg\surnameend}
  (\bibinfo{year}{2016}): \emph{\bibinfo{title}{Ssd: Single shot multibox
  detector}}.
\newblock In: {\sl \bibinfo{booktitle}{European conference on computer
  vision}}, \bibinfo{organization}{Springer}, pp. \bibinfo{pages}{21--37},
  \doi{10.1007/978-3-319-46448-0_2}.

\bibitemdeclare{inproceedings}{lou2012intelligible}
\bibitem{lou2012intelligible}
\bibinfo{author}{Yin \surnamestart Lou\surnameend}, \bibinfo{author}{Rich
  \surnamestart Caruana\surnameend} \& \bibinfo{author}{Johannes \surnamestart
  Gehrke\surnameend} (\bibinfo{year}{2012}): \emph{\bibinfo{title}{Intelligible
  models for classification and regression}}.
\newblock In: {\sl \bibinfo{booktitle}{Proceedings of the 18th ACM SIGKDD
  international conference on Knowledge discovery and data mining}}, pp.
  \bibinfo{pages}{150--158}, \doi{10.1145/2339530.2339556}.

\bibitemdeclare{article}{maaten2008visualizing}
\bibitem{maaten2008visualizing}
\bibinfo{author}{Laurens van~der \surnamestart Maaten\surnameend} \&
  \bibinfo{author}{Geoffrey \surnamestart Hinton\surnameend}
  (\bibinfo{year}{2008}): \emph{\bibinfo{title}{Visualizing data using t-SNE}}.
\newblock {\sl \bibinfo{journal}{Journal of machine learning research}}
  \bibinfo{volume}{9}(\bibinfo{number}{Nov}), pp. \bibinfo{pages}{2579--2605}.

\bibitemdeclare{inproceedings}{mangin:hal-00652346}
\bibitem{mangin:hal-00652346}
\bibinfo{author}{Olivier \surnamestart Mangin\surnameend} \&
  \bibinfo{author}{Pierre-Yves \surnamestart Oudeyer\surnameend}
  (\bibinfo{year}{2011}): \emph{\bibinfo{title}{{Unsupervised learning of
  simultaneous motor primitives through imitation}}}.
\newblock In: {\sl \bibinfo{booktitle}{{IEEE ICDL-EPIROB 2011}}},
  \bibinfo{address}{Frankfurt, Germany}.
\newblock \urlprefix\url{https://hal.archives-ouvertes.fr/hal-00652346}.

\bibitemdeclare{book}{marcus2018algebraic}
\bibitem{marcus2018algebraic}
\bibinfo{author}{Gary~F \surnamestart Marcus\surnameend}
  (\bibinfo{year}{2018}): \emph{\bibinfo{title}{The algebraic mind: Integrating
  connectionism and cognitive science}}.
\newblock \bibinfo{publisher}{MIT press}, \doi{10.7551/mitpress/1187.001.0001}.

\bibitemdeclare{book}{murray1997multiple}
\bibitem{murray1997multiple}
\bibinfo{author}{Roderick \surnamestart Murray-Smith\surnameend} \&
  \bibinfo{author}{T~\surnamestart Johansen\surnameend} (\bibinfo{year}{1997}):
  \emph{\bibinfo{title}{Multiple model approaches to nonlinear modelling and
  control}}.
\newblock \bibinfo{publisher}{CRC press}.

\bibitemdeclare{inproceedings}{ng2000algorithms}
\bibitem{ng2000algorithms}
\bibinfo{author}{Andrew~Y \surnamestart Ng\surnameend},
  \bibinfo{author}{Stuart~J \surnamestart Russell\surnameend} et~al.
  (\bibinfo{year}{2000}): \emph{\bibinfo{title}{Algorithms for inverse
  reinforcement learning.}}
\newblock In: {\sl \bibinfo{booktitle}{Icml}}, \bibinfo{volume}{1},
  p.~\bibinfo{pages}{2}.

\bibitemdeclare{inproceedings}{pairet2019digital}
\bibitem{pairet2019digital}
\bibinfo{author}{{\`E}ric \surnamestart Pairet\surnameend},
  \bibinfo{author}{Paola \surnamestart Ard{\'o}n\surnameend},
  \bibinfo{author}{Xingkun \surnamestart Liu\surnameend},
  \bibinfo{author}{Jos{\'e} \surnamestart Lopes\surnameend},
  \bibinfo{author}{Helen \surnamestart Hastie\surnameend} \&
  \bibinfo{author}{Katrin~S \surnamestart Lohan\surnameend}
  (\bibinfo{year}{2019}): \emph{\bibinfo{title}{A Digital Twin for Human-Robot
  Interaction}}.
\newblock In: {\sl \bibinfo{booktitle}{2019 14th ACM/IEEE International
  Conference on Human-Robot Interaction (HRI)}}, \bibinfo{organization}{IEEE},
  pp. \bibinfo{pages}{372--372}, \doi{10.1109/HRI.2019.8673015}.

\bibitemdeclare{inproceedings}{pastor2009learning}
\bibitem{pastor2009learning}
\bibinfo{author}{Peter \surnamestart Pastor\surnameend}, \bibinfo{author}{Heiko
  \surnamestart Hoffmann\surnameend}, \bibinfo{author}{Tamim \surnamestart
  Asfour\surnameend} \& \bibinfo{author}{Stefan \surnamestart
  Schaal\surnameend} (\bibinfo{year}{2009}): \emph{\bibinfo{title}{Learning and
  generalization of motor skills by learning from demonstration}}.
\newblock In: {\sl \bibinfo{booktitle}{2009 IEEE International Conference on
  Robotics and Automation}}, \bibinfo{organization}{IEEE}, pp.
  \bibinfo{pages}{763--768}, \doi{10.1109/ROBOT.2009.5152385}.

\bibitemdeclare{article}{penkov2017using}
\bibitem{penkov2017using}
\bibinfo{author}{Svetlin \surnamestart Penkov\surnameend} \&
  \bibinfo{author}{Subramanian \surnamestart Ramamoorthy\surnameend}
  (\bibinfo{year}{2017}): \emph{\bibinfo{title}{Using program induction to
  interpret transition system dynamics}}.
\newblock {\sl \bibinfo{journal}{arXiv preprint arXiv:1708.00376}}.

\bibitemdeclare{inproceedings}{pinto2016supersizing}
\bibitem{pinto2016supersizing}
\bibinfo{author}{Lerrel \surnamestart Pinto\surnameend} \&
  \bibinfo{author}{Abhinav \surnamestart Gupta\surnameend}
  (\bibinfo{year}{2016}): \emph{\bibinfo{title}{Supersizing self-supervision:
  Learning to grasp from 50k tries and 700 robot hours}}.
\newblock In: {\sl \bibinfo{booktitle}{2016 IEEE international conference on
  robotics and automation (ICRA)}}, \bibinfo{organization}{IEEE}, pp.
  \bibinfo{pages}{3406--3413}, \doi{10.1109/ICRA.2016.7487517}.

\bibitemdeclare{inproceedings}{qi2017pointnet}
\bibitem{qi2017pointnet}
\bibinfo{author}{Charles~R \surnamestart Qi\surnameend}, \bibinfo{author}{Hao
  \surnamestart Su\surnameend}, \bibinfo{author}{Kaichun \surnamestart
  Mo\surnameend} \& \bibinfo{author}{Leonidas~J \surnamestart
  Guibas\surnameend} (\bibinfo{year}{2017}): \emph{\bibinfo{title}{Pointnet:
  Deep learning on point sets for 3d classification and segmentation}}.
\newblock In: {\sl \bibinfo{booktitle}{Proceedings of the IEEE Conference on
  Computer Vision and Pattern Recognition}}, pp. \bibinfo{pages}{652--660}.

\bibitemdeclare{article}{rajeswaran2017learning}
\bibitem{rajeswaran2017learning}
\bibinfo{author}{Aravind \surnamestart Rajeswaran\surnameend},
  \bibinfo{author}{Vikash \surnamestart Kumar\surnameend},
  \bibinfo{author}{Abhishek \surnamestart Gupta\surnameend},
  \bibinfo{author}{Giulia \surnamestart Vezzani\surnameend},
  \bibinfo{author}{John \surnamestart Schulman\surnameend},
  \bibinfo{author}{Emanuel \surnamestart Todorov\surnameend} \&
  \bibinfo{author}{Sergey \surnamestart Levine\surnameend}
  (\bibinfo{year}{2017}): \emph{\bibinfo{title}{Learning complex dexterous
  manipulation with deep reinforcement learning and demonstrations}}.
\newblock {\sl \bibinfo{journal}{arXiv preprint arXiv:1709.10087}},
  \doi{10.15607/RSS.2018.XIV.049}.

\bibitemdeclare{article}{redmon2018yolov3}
\bibitem{redmon2018yolov3}
\bibinfo{author}{Joseph \surnamestart Redmon\surnameend} \&
  \bibinfo{author}{Ali \surnamestart Farhadi\surnameend}
  (\bibinfo{year}{2018}): \emph{\bibinfo{title}{Yolov3: An incremental
  improvement}}.
\newblock {\sl \bibinfo{journal}{arXiv preprint arXiv:1804.02767}}.

\bibitemdeclare{inproceedings}{ren2015faster}
\bibitem{ren2015faster}
\bibinfo{author}{Shaoqing \surnamestart Ren\surnameend},
  \bibinfo{author}{Kaiming \surnamestart He\surnameend}, \bibinfo{author}{Ross
  \surnamestart Girshick\surnameend} \& \bibinfo{author}{Jian \surnamestart
  Sun\surnameend} (\bibinfo{year}{2015}): \emph{\bibinfo{title}{Faster r-cnn:
  Towards real-time object detection with region proposal networks}}.
\newblock In: {\sl \bibinfo{booktitle}{Advances in neural information
  processing systems}}, pp. \bibinfo{pages}{91--99},
  \doi{10.1109/TPAMI.2016.2577031}.

\bibitemdeclare{inproceedings}{ribeiro2016should}
\bibitem{ribeiro2016should}
\bibinfo{author}{Marco~Tulio \surnamestart Ribeiro\surnameend},
  \bibinfo{author}{Sameer \surnamestart Singh\surnameend} \&
  \bibinfo{author}{Carlos \surnamestart Guestrin\surnameend}
  (\bibinfo{year}{2016}): \emph{\bibinfo{title}{" Why should I trust you?"
  Explaining the predictions of any classifier}}.
\newblock In: {\sl \bibinfo{booktitle}{Proceedings of the 22nd ACM SIGKDD
  international conference on knowledge discovery and data mining}}, pp.
  \bibinfo{pages}{1135--1144}, \doi{10.18653/v1/N16-3020}.

\bibitemdeclare{misc}{husky}
\bibitem{husky}
\bibinfo{author}{Clearpath \surnamestart Robotics\surnameend}
  (\bibinfo{year}{2014}): \emph{\bibinfo{title}{Husky, unmanned ground
  vehicle}}.
\newblock
  \urlprefix\url{https://clearpathrobotics.com/husky-unmanned-ground-vehicle-robot}.

\bibitemdeclare{article}{samek2017explainable}
\bibitem{samek2017explainable}
\bibinfo{author}{Wojciech \surnamestart Samek\surnameend},
  \bibinfo{author}{Thomas \surnamestart Wiegand\surnameend} \&
  \bibinfo{author}{Klaus-Robert \surnamestart M{\"u}ller\surnameend}
  (\bibinfo{year}{2017}): \emph{\bibinfo{title}{Explainable artificial
  intelligence: Understanding, visualizing and interpreting deep learning
  models}}.
\newblock {\sl \bibinfo{journal}{arXiv preprint arXiv:1708.08296}}.

\bibitemdeclare{inproceedings}{sato2001rule}
\bibitem{sato2001rule}
\bibinfo{author}{Makoto \surnamestart Sato\surnameend} \&
  \bibinfo{author}{Hiroshi \surnamestart Tsukimoto\surnameend}
  (\bibinfo{year}{2001}): \emph{\bibinfo{title}{Rule extraction from neural
  networks via decision tree induction}}.
\newblock In: {\sl \bibinfo{booktitle}{IJCNN'01. International Joint Conference
  on Neural Networks. Proceedings (Cat. No. 01CH37222)}}, \bibinfo{volume}{3},
  \bibinfo{organization}{IEEE}, pp. \bibinfo{pages}{1870--1875},
  \doi{10.1109/IJCNN.2001.938448}.

\bibitemdeclare{incollection}{schaal2006dynamic}
\bibitem{schaal2006dynamic}
\bibinfo{author}{Stefan \surnamestart Schaal\surnameend}
  (\bibinfo{year}{2006}): \emph{\bibinfo{title}{Dynamic movement primitives-a
  framework for motor control in humans and humanoid robotics}}.
\newblock In: {\sl \bibinfo{booktitle}{Adaptive motion of animals and
  machines}}, \bibinfo{publisher}{Springer}, pp. \bibinfo{pages}{261--280},
  \doi{10.1007/4-431-31381-8_23}.

\bibitemdeclare{inproceedings}{schaal2005learning}
\bibitem{schaal2005learning}
\bibinfo{author}{Stefan \surnamestart Schaal\surnameend}, \bibinfo{author}{Jan
  \surnamestart Peters\surnameend}, \bibinfo{author}{Jun \surnamestart
  Nakanishi\surnameend} \& \bibinfo{author}{Auke \surnamestart
  Ijspeert\surnameend} (\bibinfo{year}{2005}): \emph{\bibinfo{title}{Learning
  movement primitives}}.
\newblock In: {\sl \bibinfo{booktitle}{Robotics research. the eleventh
  international symposium}}, \bibinfo{organization}{Springer}, pp.
  \bibinfo{pages}{561--572}, \doi{10.1007/11008941_60}.

\bibitemdeclare{inproceedings}{selvaraju2017grad}
\bibitem{selvaraju2017grad}
\bibinfo{author}{Ramprasaath~R \surnamestart Selvaraju\surnameend},
  \bibinfo{author}{Michael \surnamestart Cogswell\surnameend},
  \bibinfo{author}{Abhishek \surnamestart Das\surnameend},
  \bibinfo{author}{Ramakrishna \surnamestart Vedantam\surnameend},
  \bibinfo{author}{Devi \surnamestart Parikh\surnameend} \&
  \bibinfo{author}{Dhruv \surnamestart Batra\surnameend}
  (\bibinfo{year}{2017}): \emph{\bibinfo{title}{Grad-cam: Visual explanations
  from deep networks via gradient-based localization}}.
\newblock In: {\sl \bibinfo{booktitle}{Proceedings of the IEEE International
  Conference on Computer Vision}}, pp. \bibinfo{pages}{618--626},
  \doi{10.1007/s11263-019-01228-7}.

\bibitemdeclare{article}{simonyan2013deep}
\bibitem{simonyan2013deep}
\bibinfo{author}{Karen \surnamestart Simonyan\surnameend},
  \bibinfo{author}{Andrea \surnamestart Vedaldi\surnameend} \&
  \bibinfo{author}{Andrew \surnamestart Zisserman\surnameend}
  (\bibinfo{year}{2013}): \emph{\bibinfo{title}{Deep inside convolutional
  networks: Visualising image classification models and saliency maps}}.
\newblock {\sl \bibinfo{journal}{arXiv preprint arXiv:1312.6034}}.

\bibitemdeclare{article}{springenberg2015unsupervised}
\bibitem{springenberg2015unsupervised}
\bibinfo{author}{Jost~Tobias \surnamestart Springenberg\surnameend}
  (\bibinfo{year}{2015}): \emph{\bibinfo{title}{Unsupervised and
  semi-supervised learning with categorical generative adversarial networks}}.
\newblock {\sl \bibinfo{journal}{arXiv preprint arXiv:1511.06390}}.

\bibitemdeclare{incollection}{sweller2011cognitive}
\bibitem{sweller2011cognitive}
\bibinfo{author}{John \surnamestart Sweller\surnameend} (\bibinfo{year}{2011}):
  \emph{\bibinfo{title}{Cognitive load theory}}.
\newblock In: {\sl \bibinfo{booktitle}{Psychology of learning and motivation}},
  \bibinfo{volume}{55}, \bibinfo{publisher}{Elsevier}, pp.
  \bibinfo{pages}{37--76}, \doi{10.1007/978-1-4419-1428-6_446}.

\bibitemdeclare{article}{van2019interpretable}
\bibitem{van2019interpretable}
\bibinfo{author}{Arnaud \surnamestart Van~Looveren\surnameend} \&
  \bibinfo{author}{Janis \surnamestart Klaise\surnameend}
  (\bibinfo{year}{2019}): \emph{\bibinfo{title}{Interpretable counterfactual
  explanations guided by prototypes}}.
\newblock {\sl \bibinfo{journal}{arXiv preprint arXiv:1907.02584}}.

\bibitemdeclare{article}{wachter2017counterfactual}
\bibitem{wachter2017counterfactual}
\bibinfo{author}{Sandra \surnamestart Wachter\surnameend},
  \bibinfo{author}{Brent \surnamestart Mittelstadt\surnameend} \&
  \bibinfo{author}{Chris \surnamestart Russell\surnameend}
  (\bibinfo{year}{2017}): \emph{\bibinfo{title}{Counterfactual Explanations
  without Opening the Black Box: Automated Decisions and the GPDR}}.
\newblock {\sl \bibinfo{journal}{Harv. JL \& Tech.}} \bibinfo{volume}{31}, p.
  \bibinfo{pages}{841}, \doi{10.2139/ssrn.3063289}.

\bibitemdeclare{inproceedings}{wang2016dueling}
\bibitem{wang2016dueling}
\bibinfo{author}{Ziyu \surnamestart Wang\surnameend}, \bibinfo{author}{Tom
  \surnamestart Schaul\surnameend}, \bibinfo{author}{Matteo \surnamestart
  Hessel\surnameend}, \bibinfo{author}{Hado \surnamestart Hasselt\surnameend},
  \bibinfo{author}{Marc \surnamestart Lanctot\surnameend} \&
  \bibinfo{author}{Nando \surnamestart Freitas\surnameend}
  (\bibinfo{year}{2016}): \emph{\bibinfo{title}{Dueling network architectures
  for deep reinforcement learning}}.
\newblock In: {\sl \bibinfo{booktitle}{International conference on machine
  learning}}, pp. \bibinfo{pages}{1995--2003}.

\bibitemdeclare{article}{842154}
\bibitem{842154}
\bibinfo{author}{\surnamestart {Yaochu Jin}\surnameend} (\bibinfo{year}{2000}):
  \emph{\bibinfo{title}{Fuzzy modeling of high-dimensional systems: complexity
  reduction and interpretability improvement}}.
\newblock {\sl \bibinfo{journal}{IEEE Transactions on Fuzzy Systems}}
  \bibinfo{volume}{8}(\bibinfo{number}{2}), pp. \bibinfo{pages}{212--221},
  \doi{10.1109/91.842154}.

\bibitemdeclare{article}{yu2018one}
\bibitem{yu2018one}
\bibinfo{author}{Tianhe \surnamestart Yu\surnameend}, \bibinfo{author}{Chelsea
  \surnamestart Finn\surnameend}, \bibinfo{author}{Annie \surnamestart
  Xie\surnameend}, \bibinfo{author}{Sudeep \surnamestart Dasari\surnameend},
  \bibinfo{author}{Tianhao \surnamestart Zhang\surnameend},
  \bibinfo{author}{Pieter \surnamestart Abbeel\surnameend} \&
  \bibinfo{author}{Sergey \surnamestart Levine\surnameend}
  (\bibinfo{year}{2018}): \emph{\bibinfo{title}{One-shot imitation from
  observing humans via domain-adaptive meta-learning}}.
\newblock {\sl \bibinfo{journal}{arXiv preprint arXiv:1802.01557}},
  \doi{10.15607/RSS.2018.XIV.002}.

\bibitemdeclare{inproceedings}{zeiler2014visualizing}
\bibitem{zeiler2014visualizing}
\bibinfo{author}{Matthew~D \surnamestart Zeiler\surnameend} \&
  \bibinfo{author}{Rob \surnamestart Fergus\surnameend} (\bibinfo{year}{2014}):
  \emph{\bibinfo{title}{Visualizing and understanding convolutional networks}}.
\newblock In: {\sl \bibinfo{booktitle}{European conference on computer
  vision}}, \bibinfo{organization}{Springer}, pp. \bibinfo{pages}{818--833},
  \doi{10.1007/978-3-319-10590-1_53}.

\bibitemdeclare{techreport}{zhu2005semi}
\bibitem{zhu2005semi}
\bibinfo{author}{Xiaojin~Jerry \surnamestart Zhu\surnameend}
  (\bibinfo{year}{2005}): \emph{\bibinfo{title}{Semi-supervised learning
  literature survey}}.
\newblock \bibinfo{type}{Technical Report}, \bibinfo{institution}{University of
  Wisconsin-Madison Department of Computer Sciences}.

\bibitemdeclare{article}{zhu2018reinforcement}
\bibitem{zhu2018reinforcement}
\bibinfo{author}{Yuke \surnamestart Zhu\surnameend}, \bibinfo{author}{Ziyu
  \surnamestart Wang\surnameend}, \bibinfo{author}{Josh \surnamestart
  Merel\surnameend}, \bibinfo{author}{Andrei \surnamestart Rusu\surnameend},
  \bibinfo{author}{Tom \surnamestart Erez\surnameend}, \bibinfo{author}{Serkan
  \surnamestart Cabi\surnameend}, \bibinfo{author}{Saran \surnamestart
  Tunyasuvunakool\surnameend}, \bibinfo{author}{J{\'a}nos \surnamestart
  Kram{\'a}r\surnameend}, \bibinfo{author}{Raia \surnamestart
  Hadsell\surnameend}, \bibinfo{author}{Nando \surnamestart
  de~Freitas\surnameend} et~al. (\bibinfo{year}{2018}):
  \emph{\bibinfo{title}{Reinforcement and imitation learning for diverse
  visuomotor skills}}.
\newblock {\sl \bibinfo{journal}{arXiv preprint arXiv:1802.09564}},
  \doi{10.15607/RSS.2018.XIV.009}.

\bibitemdeclare{inproceedings}{zilke2016deepred}
\bibitem{zilke2016deepred}
\bibinfo{author}{Jan~Ruben \surnamestart Zilke\surnameend},
  \bibinfo{author}{Eneldo~Loza \surnamestart Menc{\'\i}a\surnameend} \&
  \bibinfo{author}{Frederik \surnamestart Janssen\surnameend}
  (\bibinfo{year}{2016}): \emph{\bibinfo{title}{Deepred--rule extraction from
  deep neural networks}}.
\newblock In: {\sl \bibinfo{booktitle}{International Conference on Discovery
  Science}}, \bibinfo{organization}{Springer}, pp. \bibinfo{pages}{457--473},
  \doi{10.1007/978-3-319-46307-0_29}.

\end{thebibliography}
%\bibliographystyle{IEEEtran}
%\bibliography{root}

\appendix
\section{Inferred programs for extended scenarios}
\label{app:programs}

Programs for the extended scenarios presented in Fig.~\ref{fig:newscenarios}.
The command \emph{execute(goal_label)} calls the controller to arrive from the previous goal to the one defined as parameter. If there is no previous goal, the origin is assumed. Goal number correspond to position as presented in Fig.~\ref{fig:seq-1-2-a}. The labels with cardinal direction (1W, 1E, 2N, 2S, 3W, 3E, 4N and 4S) represent the goals in the middle of the halls.

\begin{minipage}{.96\textwidth}
  \begin{lstlisting}[language=Python, caption={Extended scenario (a).
}, label=lis:program4a]
def program():
	execute(1)
	execute(1W)
	execute(2)
	execute(2N)
	execute(3)
	execute(3E)
	execute(4)
	execute(4S)
	execute(1)
	execute(4N)
	execute(4)
	execute(3W)
	execute(3)
	execute(2S)
	execute(2)
	execute(1E)
	execute(1)
	return
\end{lstlisting}
\end{minipage}

\begin{minipage}{.96\textwidth}
  \begin{lstlisting}[language=Python, caption={Extended scenario (b).
}, label=lis:program4b]
def program():
	controller_list = [1,2]
	count = 0
	for k in range(len(controller_list)*2-1):
		execute(controller_list[count])
		if k >= len(controller_list)-1:
			count = count-1
		else:
			count = count+1
	controller_list = [4,1,2,3]
	count = 0
	for k in range(len(controller_list)*2-1):
		execute(controller_list[count])
		if k >= len(controller_list)-1:
			count = count-1
		else:
			count = count+1
	execute(3)
	execute(4)
	execute(1)
	return
\end{lstlisting}
\end{minipage}

\begin{minipage}{.96\textwidth}
\begin{lstlisting}[language=Python, caption={Extended scenario (c).
}, label=lis:program4c]
def program():
	for j in range(2):
		controller_list = [1,2,3]
		count = 0
		for k in range(len(controller_list)*2-1):
			execute(controller_list[count])
			if k >= len(controller_list)-1:
				count = count-1
			else:
				count = count+1
		controller_list = [4,3]
		count = 0
		for k in range(len(controller_list)*2-1):
			execute(controller_list[count])
			if k >= len(controller_list)-1:
				count = count-1
			else:
				count = count+1
	execute(1)
	return
\end{lstlisting}
\end{minipage}
\end{document}